\documentclass[10pt, letter, onecolumn]{arxiv}

\usepackage{kantlipsum, lipsum}
\usepackage{dm-colors}
\usepackage{amsmath}
\usepackage{pstricks, pst-node}
\usepackage{verbatim}
\usepackage{multirow}
\usepackage{scalerel}
\usepackage{booktabs}
\usepackage{enumitem}
\usepackage{xspace}
\usepackage{bm}
\usepackage{bbm}
\usepackage{mathtools}
\usepackage{soul}
\usepackage{epsfig}
\usepackage{graphicx}
\usepackage{tcolorbox}
\usepackage{subcaption}
\usepackage{amssymb}
\usepackage{colortbl}
\usepackage{csquotes}
\usepackage{etex}
\usepackage{setspace}
\usepackage{svg}
\usepackage{colortbl}
\usepackage{tabularx,ragged2e}
\usepackage{placeins}
\usepackage{afterpage}
\usepackage[symbol]{footmisc}
\usepackage[bibstyle=nature,citestyle=numeric-comp,%
            natbib=true,backend=biber,maxbibnames=99,%
            giveninits=false,sorting=none]{biblatex}
\usepackage{nameref}
\usepackage{varioref}
\usepackage[pagebackref=false,breaklinks=false,%
            colorlinks=true,bookmarks=true,citecolor=ourdarkblue,%
            urlcolor=ourdarkblue,linkcolor=ourdarkblue]{hyperref}
\usepackage[noabbrev,capitalize]{cleveref}

\usepackage{etoc}
\usepackage{lineno}
\graphicspath{{figures/}}

\usepackage{listings}
\DeclareFixedFont{\ttb}{T1}{txtt}{bx}{n}{9} 
\DeclareFixedFont{\ttm}{T1}{txtt}{m}{n}{9}  
\usepackage{color}
\definecolor{deepblue}{rgb}{0,0,0.5}
\definecolor{deepred}{rgb}{0.6,0,0}
\definecolor{deepgreen}{rgb}{0,0.5,0}
\newcommand\pythonstyle{\lstset{
language=Python,
basicstyle=\footnotesize,
morekeywords={self},              
keywordstyle=\color{deepblue},
emph={DDx,MxPlan,ChainedDDx,ChainedMxPlan,__init__,PatientSummary,Message,Dialogue,Outline,Outlines,Diagnosis,Intervention,Likelihood,Literal,Citation,ReasoningSteps,PlannerOutput,GeneratedFields,SearchQueries,Annotated, AutoevalMxRating,AorB,Critique,GoodBad,ActionItem},          
emphstyle=\color{deepred},    
stringstyle=\color{deepgreen},
commentstyle=\color{deepgreen},
showstringspaces=false
}}
\lstnewenvironment{python}[1][]
{
\pythonstyle
\lstset{#1}
}
{}

\newcommand\pythoninline[1]{{\pythonstyle\lstinline!#1!}}

\newtcolorbox{promptbox2}[1]{ 
    colback=black!5!white,
    colframe=black!60!white,
    arc=1mm, 
    boxrule=1pt, 
    bottomrule=2pt,
    title=#1, 
    fonttitle=\bfseries,
    fontupper=\fontfamily{lmtt},
    fontlower=\fontfamily{lmtt},
    capture=minipage 
}

\def\numpcps{21\xspace}
\def\numpatientactors{21\xspace}
\def\numspecialists{10\xspace}

\title{{Towards Conversational AI for\\Disease Management}}

\author[$\ast$,$\ddagger$,1]{Anil Palepu}
\author[$\ast$,$\ddagger$,2]{Valentin Liévin}
\author[1]{\\Wei-Hung Weng}
\author[2]{Khaled Saab}
\author[2]{David Stutz}
\author[2]{Yong Cheng}
\author[1]{\\Kavita Kulkarni}
\author[2]{S. Sara Mahdavi}
\author[2]{Joëlle Barral}
\author[1]{Dale R. Webster}
\author[1]{\\Katherine Chou}
\author[1]{Avinatan Hassidim}
\author[1]{Yossi Matias}
\author[1]{James Manyika}
\author[2]{\\Ryutaro Tanno}
\author[1]{Vivek Natarajan}
\author[1]{Adam Rodman}
\author[2]{Tao Tu}
\author[$\dagger$,$\ddagger$,1]{\\Alan Karthikesalingam}
\author[$\dagger$,$\ddagger$,1]{Mike Schaekermann}

\affil[1]{Google Research, }
\affil[2]{Google DeepMind}

\renewcommand{\correspondingauthor}[1]{$\ast$~Equal contributions. %
                                       $\dagger$~Equal leadership. \\%
                                       $\ddagger$~Corresponding authors: \{apalepu, vlievin, alankarthi, mikeshake\}@google.com }

\begin{document}

\begin{refsection}

\begin{abstract}

While large language models (LLMs) have shown promise in diagnostic dialogue, their capabilities for effective management reasoning---including disease progression, therapeutic response, and safe medication prescription---remain under-explored. We advance the previously demonstrated diagnostic capabilities of the Articulate Medical Intelligence Explorer (AMIE) through a new LLM-based agentic system optimised for clinical management and dialogue, incorporating reasoning over the evolution of disease and multiple patient visit encounters, response to therapy, and professional competence in medication prescription. To ground its reasoning in authoritative clinical knowledge, AMIE leverages Gemini's long-context capabilities, combining in-context retrieval with structured reasoning to align its output with relevant and up-to-date clinical practice guidelines and drug formularies. In a randomized, blinded virtual Objective Structured Clinical Examination (OSCE) study, AMIE was compared to 21 primary care physicians (PCPs) across 100 multi-visit case scenarios designed to reflect UK NICE Guidance and BMJ Best Practice guidelines. AMIE was non-inferior to PCPs in management reasoning as assessed by specialist physicians and scored better in both preciseness of treatments and investigations, and in its alignment with and grounding of management plans in clinical guidelines. To benchmark medication reasoning, we developed RxQA, a multiple-choice question benchmark derived from two national drug formularies (US, UK) and validated by board-certified pharmacists. While AMIE and PCPs both benefited from the ability to access external drug information, AMIE outperformed PCPs on higher difficulty questions. While further research would be needed before real-world translation, AMIE's strong performance across evaluations marks a significant step towards conversational AI as a tool in disease management.

\end{abstract}

\maketitle


\afterpage{
\begin{figure}[]
    \centering
    \includegraphics[width=1\textwidth,height=\textheight,keepaspectratio]{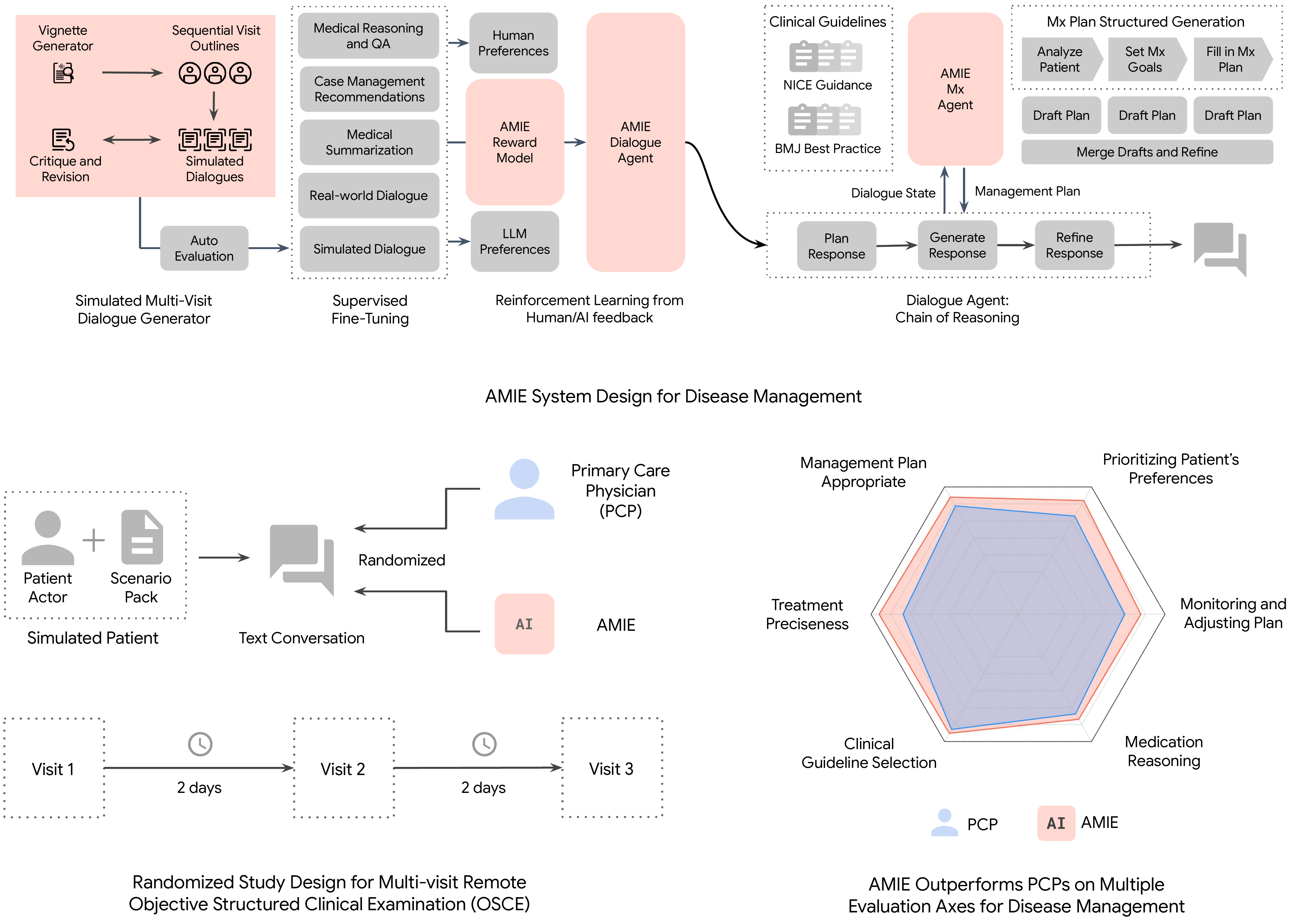}
    \vspace{0.4cm}
    \caption{\textbf{Overview of contributions.} AMIE's capabilities were advanced towards the management of disease over the course of multiple visits and in concordance with authoritative knowledge in the form of clinical guidelines. AMIE leverages an agentic system incorporating a Dialogue Agent for conversational interaction and a Management Reasoning (Mx) Agent for evidence-based reasoning and management plan generation. A simulated dialogue environment enables refinement of AMIE's capabilities across diverse medical contexts. During online inference, AMIE utilizes Gemini's long-context capabilities to reason over multiple clinical guidelines and generate a comprehensive management plan. AMIE's performance is evaluated through a multi-visit remote OSCE study comparing it to primary care physicians.}
    \label{fig:contributions}
\end{figure}
}

\section{Introduction}
\label{sec:introduction}

There has been rapid advancement in the promise demonstrated by large language models (LLMs) for clinical applications, with studies demonstrating not only the ability to render accurate differential diagnoses when given curated patient information, but also for LLM-based Artificial Intelligence (AI) systems to actively collect medical history through empathetic conversations in a natural style that builds trust and rapport\cite{tu2024towards,kanjee2023accuracy,mcduff2023towards}. However, this task, often called diagnostic reasoning, is necessary but not sufficient for clinical care applications. Beyond taking a history, human physicians must select  appropriate investigations for a patient, create an acceptable care plan that takes patient preferences and system constraints into account, and consider the progression of the disease and its treatment over time, including such decisions as ``watchful waiting'', or interval follow up. This cognitive process, known as management reasoning, requires complex synthesis of relevant clinical guidelines, new and evolving evidence in medical literature, and local patient-specific context~\cite{Cook2018-xs}. 

While the study of diagnostic reasoning has a long history with many validated measures for both human and AI evaluation \cite{Monteiro2020-jk},  there has been comparatively little investigation into the evaluation of management reasoning. This is largely because management reasoning shows considerable amounts of context specificity --- that is, two physicians given the exact same patient and information still might make different decisions because of differing contextual factors \cite{boyle2025context}.  Most research on management reasoning has focused on medical education; the current gold standard in assessment is the Objective Structured Clinical Evaluation (OSCE), in which a student interviews and makes management decisions on a patient actor, and is evaluated via a standardized rubric.  There have been few rigorous investigations of the capabilities of LLMs to perform management reasoning, and the studies that have been performed rely on static non-conversational settings~\cite{Goh2025-wa}.

In this paper, we detail our progress towards optimizing the capabilities of the Articulate Medical Intelligence Explorer (AMIE) -- an LLM-based research AI system with physician-like performance on conversational diagnostic tasks -- for  management reasoning over time. We developed an LLM-based agentic system optimised for clinical management and dialogue, incorporating reasoning over the longitudinal evolution of disease and multiple patient visit encounters, response to therapy, and professional competence in medication prescription. Using a randomized, blinded virtual OSCE study as an analogue, we compared AMIE to primary care physicians and evaluated their behavior on multiple domains of management reasoning.

\newpage
Our contributions are summarized in \cref{fig:contributions}:

\begin{itemize}[leftmargin=1.5em,rightmargin=0em]
\setlength\itemsep{5pt}
\item We advance the previously demonstrated diagnostic capabilities of AMIE through a new LLM-based agentic system optimised for clinical management and dialogue, incorporating reasoning over the longitudinal evolution of disease and multiple patient visit encounters, response to therapy, and professional competence in medication prescription. To ground its reasoning in authoritative clinical knowledge, AMIE leverages Gemini's long-context capabilities, combining in-context retrieval with structured reasoning to align its output with relevant and up-to-date clinical practice guidelines and drug formularies.
\item We conducted a randomized, blinded virtual OSCE study, comparing AMIE to \numpcps primary care physicians (PCPs) across 100 multi-visit case scenarios and five medical specialties, derived to entail decision-making discussed in UK NICE Guidance and BMJ Best Practice guidelines. Based on the OSCE interactions, specialist physicians and patient actors assessed AMIE and PCPs on multiple domains of management reasoning. To benchmark medication reasoning, we developed RxQA, a challenging multiple-choice question benchmark derived from two national drug formularies (US, UK) that was validated by a panel of board-certified pharmacists.
\item We demonstrate that AMIE's management reasoning capabilities overall were non-inferior to PCPs in the OSCE study, and scored better in preciseness of treatments and investigations, and in its alignment with and grounding of management plans in clinical guidelines. For medication reasoning (RxQA), while AMIE and PCPs both benefited significantly from the ability to access external drug information, AMIE outperformed PCPs on the subset rated as higher-difficulty by pharmacists.
\end{itemize}

\section{Conversational AI for Disease Management}
\label{sec:methods}

\begin{figure}[ht!]
    \centering
    \includegraphics[width=1.0\textwidth]{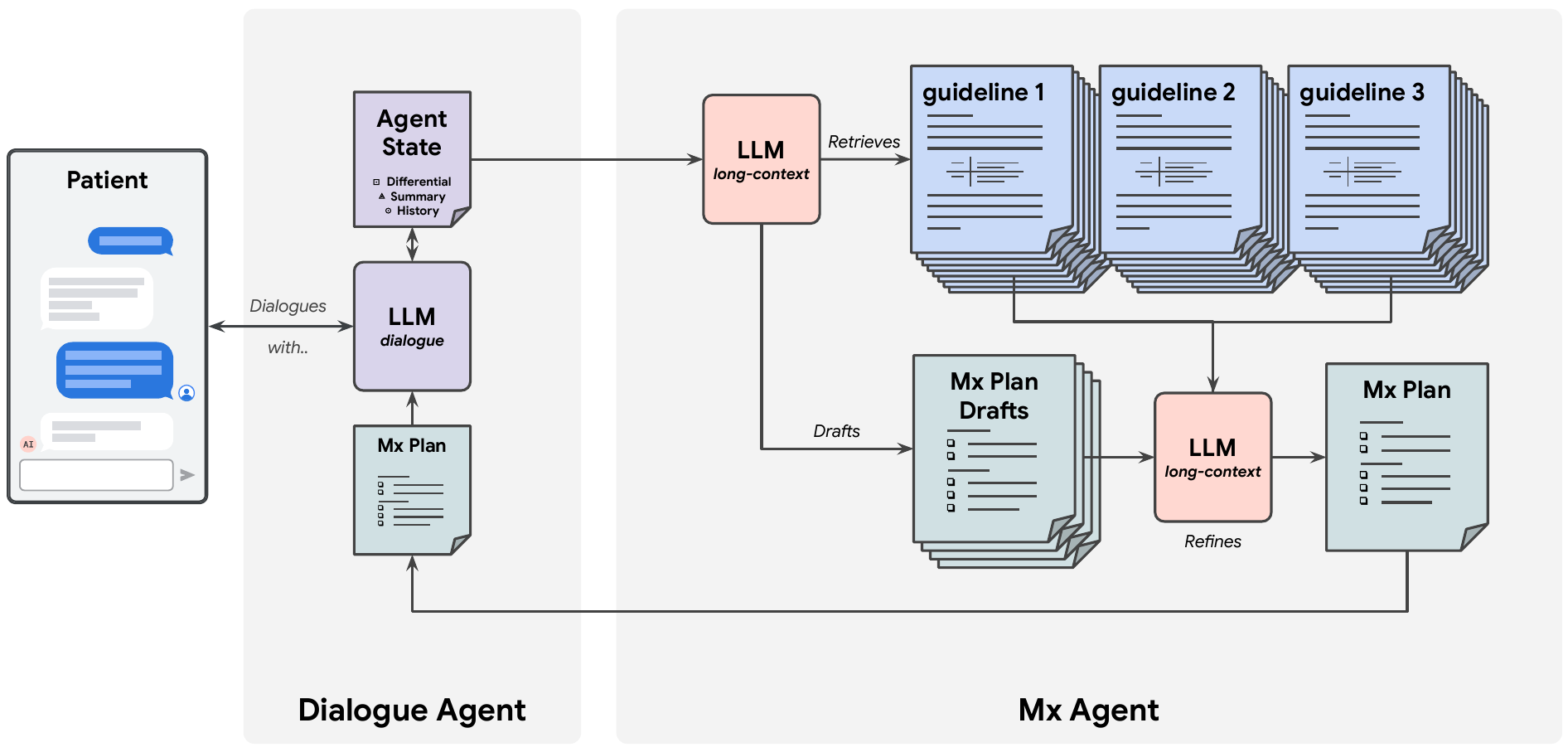}
    \vspace{0.5em}
    \caption{\textbf{System architecture.} AMIE is a symbiosis of two specialized agents: the \textit{Dialogue Agent}, whose role is to converse with the patient to collect information and communicate clinical decisions, and the \textit{Mx Agent}, whose role is to browse full-text clinical guidelines and compile tailored management plans, which are delivered to the Dialogue Agent.}
    \label{fig:system_architecture}
\end{figure}

In this work, we optimised AMIE for a setting where it would converse with a patient via synchronous text chat in real time and over the course of multiple patient visit encounters.
To reconcile the need for swift responses in synchronous conversation with the need for in-depth clinical reasoning, we employ a multi-agent system, inspired by the dual-process theory of cognition popularized by \citeauthor{kahneman2011thinking} in ``Thinking, Fast and Slow'' \cite{kahneman2011thinking} and also discussed in the context of diagnosis in \cite{ball2015improving}.  AMIE with management reasoning (Mx) capabilities is a system of two agents (\cref{fig:system_architecture}):

\begin{enumerate}
    \item  \textbf{Dialogue Agent}:  This agent engages in fast, intuitive and empathetic dialogue with the patient while maintaining a persistent conversational state across multiple visits. Details are described in \cref{sec:dialogue_agent}.
    \item \textbf{Mx Agent}: This agent plans the patient care through more extensive inference-time computation. It continuously analyzes the patient's case, reasons about clinical guidance from a corpus of authoritative clinical knowledge (e.g., clinical guidelines), and generates detailed and structured management plans. Details are described in \cref{sec:mx_agent}.
\end{enumerate}

In the analogy of dual-process theory, Dialogue Agent and Mx Agent correspond to `system 1' and `system 2' respectively.
Both agents are built upon Gemini language models~\cite{team2023gemini} and use reasoning to improve response quality. The Dialogue Agent, which builds on \citet{tu2024towards}, is fine-tuned for multi-visit medical conversations and diagnostic dialogue, while the Mx Agent is optimized for complex reasoning and long-context understanding over multi-visit patient contexts and hundreds of pages of full-text clinical guidelines. The Mx Agent is invoked as a tool by the Dialogue Agent, while the Dialogue Agent references the latest management plan delivered by the Mx Agent as an input to the conversation.

\subsection{Dialogue Agent}
\label{sec:dialogue_agent}

The Dialogue Agent serves two roles. Firstly, it acts as a conversational interface, quickly generating appropriate responses to the user's messages (\cref{methods:chain_of_reasoning}). Secondly, it orchestrates the rest of the system, internally managing the memory and state of the interaction (\cref{methods:agent_state}) as well as pulling the latest plan from the specialized Mx Agent, described in detail in \cref{sec:mx_agent}. 

\subsubsection{Post-training}\label{methods:dialogue_training}
We extended the simulated learning framework described in~\citet{tu2024towards} to optimize the LLM powering the Dialogue Agent for a multi-turn, multi-visit conversational setting and scale learning across a wide range of medical contexts and specialties (\cref{sec:datasets,methods:fine_tuning}). As in~\citet{tu2024towards}, we relied heavily on simulated doctor-patient dialogues and other clinically-relevant datasets to train our model; however, we made a few notable changes to improve its performance in this setting:

\begin{enumerate}
    \item \textbf{Base Model:} We built on top of Gemini 1.5 Flash instead of the previously used PaLM-2 \cite{anil2023palm}. We refer to the Gemini technical report for more details on the base LLM architecture \cite{geminiteam2024gemini15unlockingmultimodal}.
    \item \textbf{Multi-visit Simulated Dialogues:} We developed a new set of simulated dialogues (\cref{methods:synth_data}) to improve multi-visit reasoning. Prompts for dialogue generation and auto-evaluation are listed in \cref{appendix:dialogue_gen_prompts} and \cref{appendix:auto_eval_filtering} respectively. The final datasets used to fine-tune the base model for this study are a superset of those described in~\citet{tu2024towards}, including various medical question-answering tasks, EHR summarization tasks, real-world medical dialogues \cite{chiu2017speech}, and the simulated dialogue datasets.
    \item \textbf{RLHF/RLAIF:} Following supervised fine-tuning (SFT) (see \cref{appendix:sft}), we performed reinforcement learning with human/AI feedback (RLHF/RLAIF) \cite{ouyang2022training}, curating medically relevant tasks to further enhance the model's conversational and disease-management capabilities (details in \cref{appendix:rlhf}). Here, the reward model is trained from both human and LLM-generated pairwise preferences for dialogue responses, case management recommendations \cite{palepu2024exploring,o2024towards}, and more (\cref{appendix:rm_data}).
\end{enumerate}

\subsubsection{Chain-of-Reasoning}\label{methods:chain_of_reasoning}
During inference, the Dialogue Agent uses a sequence of model calls, or a ``chain-of-reasoning'' to generate its ultimate response to the patient. See \cref{prompts:chain_of_reasoning} for specific prompting details. These model calls are low latency; the Dialogue Agent's response can be output just a few seconds after a new patient message is received. Importantly, unlike the previous version of the AMIE system \cite{tu2024towards}, each chain-of-reasoning step in the Dialogue Agent is dependent not only on the dialogue history up to that point, but also on an internally managed state that is maintained across multiple patient visit encounters (\cref{methods:agent_state}). 

\newpage
The chain-of-reasoning steps for this agent are as follows:

\begin{enumerate}
    \item \textbf{Plan Response.}  The Dialogue Agent reasons about what it should do next, including details to inquire about, patient questions to respond to, and deciding whether to wrap up or continue the conversation.
    \item \textbf{Generate Response.}  With its output from the previous step in context, the Dialogue Agent drafts a response to the patient's last message.
    \item \textbf{Refine Response.} Prior to sending out the drafted response, the Dialogue Agent revises it against a set of criteria to ensure the response meets its quality standards.
\end{enumerate}

\subsubsection{Agent State}\label{methods:agent_state}
The agent state is a modular data structure that represents the agent's beliefs about the current and past conversations. This state is accessible to both the Dialogue Agent and the Mx Agent. State fields are updated periodically throughout the dialogue by an asynchronous background reasoning sub-routine. The state decomposes into the following fields:

\begin{enumerate}
    \item \textbf{Current Patient Summary.} The current patient summary condenses all confirmed positive/negative symptoms, demographics, medical/drug/family/social history, and more into a concise summary.
    \item \textbf{Current Differential Diagnosis.} The current differential lists what AMIE believes to be the most probable diagnosis and other plausible diagnoses that should be considered at a given point in time.
    \item \textbf{Current Management Plan.} The current management plan includes investigations to be done during the visit, investigations to be ordered after the visit, and recommended actions following the visit. Rather than being generated by the dialogue agent itself, this task is outsourced to the Mx Agent (\cref{sec:mx_agent}), which more thoroughly and carefully reasons over clinical guidelines to formulate this plan.
\end{enumerate}

The specific prompts used to update the agent state are described in \cref{prompts:agent_state}. Additional attributes of the agent state, such as ``visit number'', are also used to enable more context-specific logic, such as unique prompts for initial vs. follow-up visits.

\begin{figure}[ht!]
    \centering
    \includegraphics[width=1.0\textwidth]{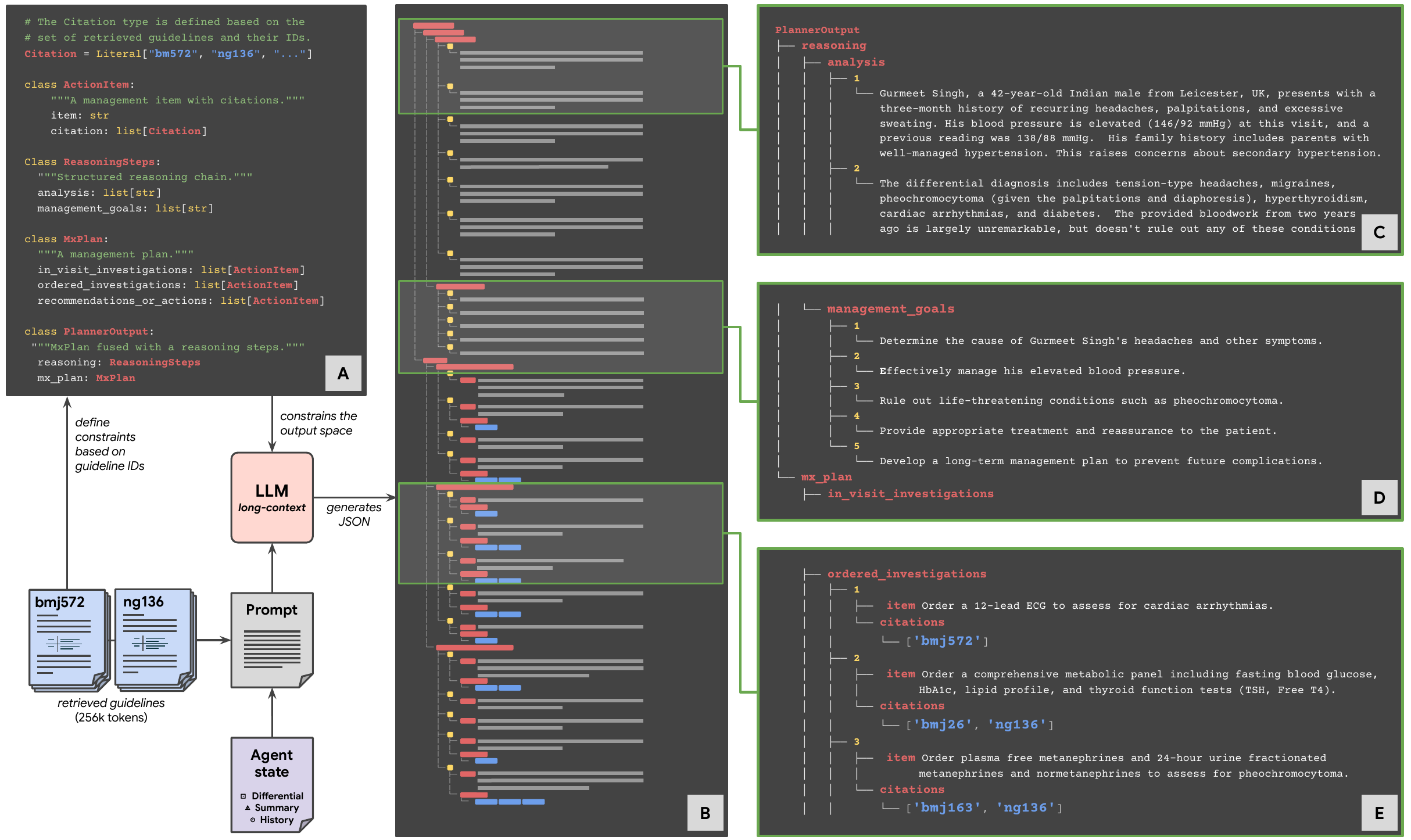}
     \vspace{0.5em}
     \caption{\textbf{Reasoning and planning under structural constraints.} Inference-time decoding constraints are applied to constrain the model output to a predefined JSON structure and sets of values. The structures are defined in Python code, generated based on the set of retrieved guidelines and automatically converted into decoding constraints. The corresponding JSON schema is appended to the prompt. Panel \textbf{A} displays the target structure represented as Python code. Panel \textbf{B} illustrates a reasoning trace generated using this structure, represented as a tree. Excerpts \textbf{C} \& \textbf{D} show parts of the reasoning steps. Excerpt \textbf{E} shows a section of the generated plan. Each plan item is annotated with generated references to the source documents through in-context reasoning over the set of retrieved guidelines.}
    \label{fig:reasoning_structures}
\end{figure}

\subsection{Mx Agent}\label{sec:mx_agent}

The goal of the Mx Agent is to plan the patient's care. Building on Gemini's state-of-the-art long-context capabilities~\cite{geminiteam2024gemini15unlockingmultimodal}, this agent synthesizes and reasons over large amounts of information---patient context across several visit encounters, as well as hundreds of pages of clinical guidelines---all at once to craft a management plan tailored to the patient's specific context. The resulting management plan is a structured set of investigations and interventions. Individual recommendations in the management plan are annotated with citations to the reference documents, such as clinical guidelines, to provide interpretability and traceability. An example is provided in \cref{appendix:mx_plan}. 

\subsubsection{Design Optimization}

Based on the expectation that management reasoning capabilities will improve with increasing test-time compute~\cite{wang2022self, lievin2022can, snell2024scaling, muennighoff2025s1,saab2024capabilities,gottweis2025aicoscientist}, the Mx Agent was designed to utilize the maximum computational power permissible under real-time user interaction constraints. For this work, we targeted a response time of no more than one minute to avoid dissociation between the two agents. We used auto-evaluation to guide our search over potential parameters (e.g., context length, prompts, inference strategy, structural constraints) for the optimal agent design. Details of this optimization process are described in \cref{appendix:mx_auto_eval}.

In the final design, the Mx Agent operates in overlapping stages (\cref{fig:system_architecture}). First, it retrieves clinical guideline documents based on generated queries and abstract embeddings. Second, it produces four draft management plans. Third, it refines and merges the drafts into a final management plan using a final generation step conditioned on retrieved documents. Plan drafting and refinement is analogous to ``ensemble refinement''~\citep{singhal2023towards}, allowing diverse management strategies to be explored concurrently. Each plan is generated by chaining long reasoning and plan construction within a single model call, enabled by structural constraints.

\subsubsection{Long-Context Reasoning}\label{methods:retrieval_long_context}

We designed the Mx Agent to tap into Gemini's long-context reasoning capabilities~\cite{geminiteam2024gemini15unlockingmultimodal,saab2024capabilities,lee2024can} rather than investing in engineering complex retrieval pipelines~\cite{gao2024retrievalaugmentedgenerationlargelanguage}, which come with inherent performance limitations~\cite{lee2024can, barnett2024sevenfailurepointsengineering} (e.g., compounding errors), technical dilemmas (e.g., choice of chunking algorithm), and often higher latency (i.e., additional communication costs). In addition to the shared agent state (which holds patient-specific information), we provide as input to the model multiple clinical guideline documents, minimally converted from their original source format to Markdown while retaining key elements of the document structure such as section headings, lists and tables, which allows preserving global semantic coherence within a single document. Note that clinical guideline documents can be extensive with a typical range of 30 to 50 pages.
Our approach enables the model to interact with in-context data at every step of the generation process for rich cross-document reasoning.

In the context of management reasoning, multi-document reasoning is often crucial for addressing complex presentations. For instance, determining the optimal treatment for a patient with comorbidities may necessitate integrating information from distinct guidelines, at minimum one for each condition. Nevertheless, tracing management recommendations back to reference documents remains an open problem, especially when using a large number of in-context sources. Rather than involving specialized tools for post-hoc citation attribution, we generate citations as an integral part of the reasoning process. Each citation task can be viewed as an instance of in-context retrieval~\cite{lee2024can}, on which long-context models like Gemini are already performing on par with specialized methods~\cite{lee2024can}.

\subsubsection{Coarse Retrieval}\label{methods:retrieval_filtering}

The corpus of clinical guidelines utilized in this study accounts for a total of 10.5 millions of tokens (across 627 documents), which exceeds Gemini's two million context window. Therefore, a preliminary retrieval step is required to coarsely filter out irrelevant documents. We built a simple retriever system using Gecko 1B text embeddings~\citep{lee2024gecko}, which we used to index all guidelines based on titles and abstracts. The abstracts are generated ahead of time, emphasizing clinically relevant features like target demographics, clinical goals and relevant conditions, symptoms, tests and interventions. Upon receiving a new patient case, the Mx agent generates up to five search queries in natural language. We used auto-evaluation (\cref{appendix:mx_auto_eval}) to determine the optimal number of context tokens (256,000) to allocate for external knowledge resources. On average, this corresponds to six clinical guideline documents which can all be synthesized and reasoned over in-context simultaneously. The prompts for query and abstract generation are provided in \cref{appendix:mx_prompt}.

\subsubsection{Structured Generation}\label{methods:mx_structured_generation}

Prompting alone does not offer guarantees around the structure used for internal reasoning steps or the generated output. This can be problematic in cases where a certain structure is expected either for reasons of interpretability and traceability (e.g., requirement of explicit citations) or to serve an interface between two components (e.g., between the Dialogue Agent and Mx Agent). To enable controllable multi-step reasoning with structural guarantees, we use decoding constraints~\citep{koo2024automata} to guide the model output using a predefined JSON schema~\cite{google-json-mode, openai-json-mode,tam2024letspeakfreelystudy}. Chaining a predefined number of reasoning steps and final plan construction together within a single model call guarantees a valid plan structure and also ensures that the predicted leaf values, such as citations, match a predefined format or range of values. Chaining multiple steps in a single model call also reduces latency as it requires encoding the documents only once.

Figure \ref{fig:reasoning_structures}.A visualizes the reasoning structure imposed by the Mx Agent. The structure is defined in Python code and automatically converted along with docstrings to JSON-equivalent decoding constraints. Auto-evaluation \cref{appendix:mx_auto_eval} was used to determine the reasoning structure for this task. We illustrate a reasoning trace generated using this structure in \cref{fig:reasoning_structures}.B. The reasoning trace consists of three components:

\begin{enumerate}
\item \textbf{Analyze Patient:} Evaluate symptoms, medical history, and overall clinical picture (\cref{fig:reasoning_structures}.C).
\item \textbf{Set Objectives:} Define high-level management goals (\cref{fig:reasoning_structures}.D).
\item \textbf{Plan and Cite:} Plan management steps and cite appropriate clinical guidelines (\cref{fig:reasoning_structures}.E).
\end{enumerate}

The model prompt for this approach is available in \cref{appendix:mx_prompt}. Several examples of reasoning traces and management plans are presented in \cref{appendix:reasoning_traces}, and further modeling details are provided in \cref{appendix:mx_auto_eval}.

\begin{figure}[ht!]
    \centering
    \includegraphics[width=\textwidth,height=\textheight,keepaspectratio]{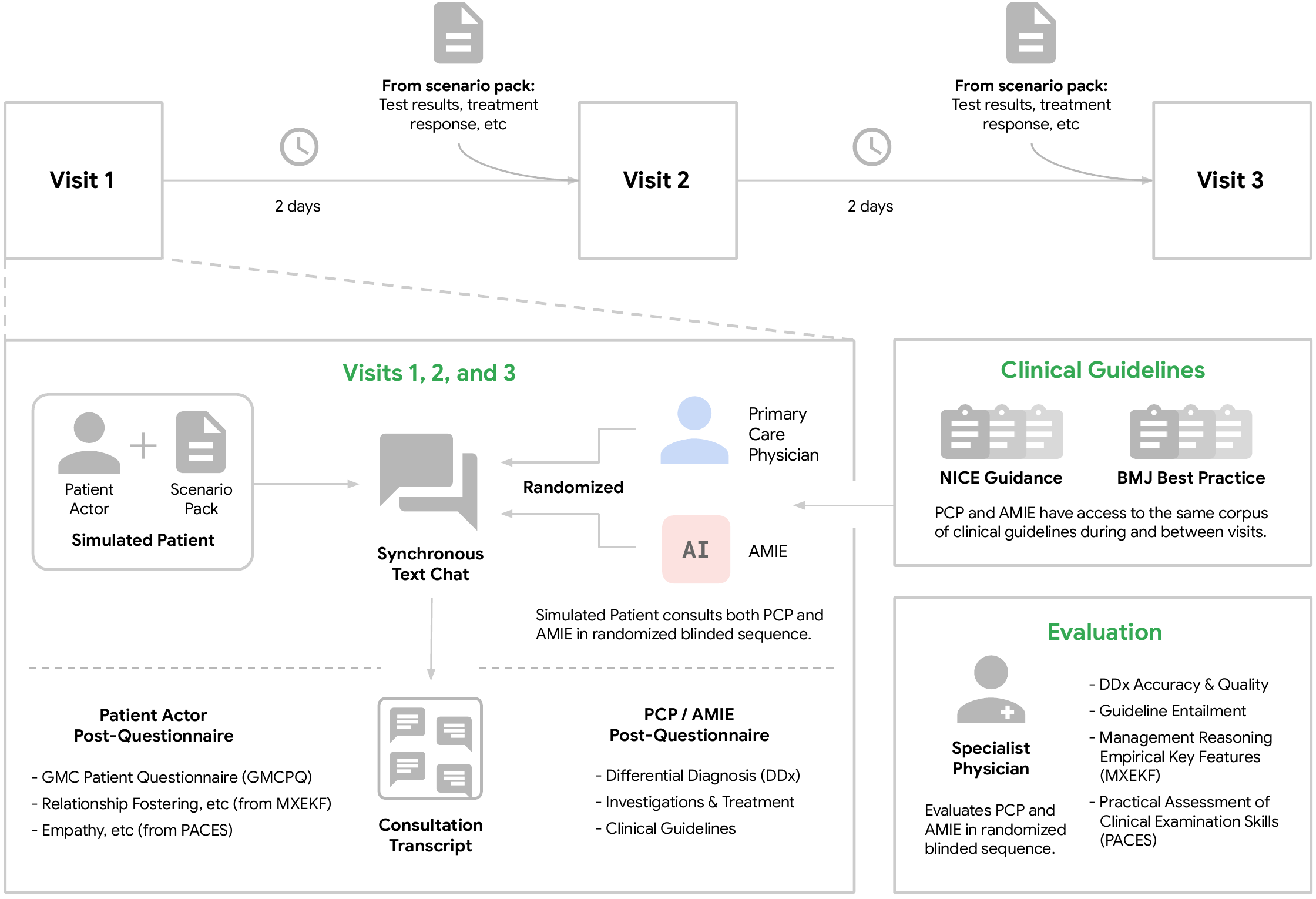}
    \vspace{0.5em}
    \caption{\textbf{Evaluation: overview of randomized study design.} A primary care physician (PCP) and AMIE perform (in a randomized order) three virtual remote OSCE visits with simulated patients via online multi-turn synchronous text chat, building upon an initial patient presentation with subsequent updates on symptoms, treatment responses, and test results.  Both the PCP and AMIE have access to a corpus of clinical guidelines. After each visit, the PCP and AMIE complete a post-questionnaire, and both are evaluated by patient actors and specialist physicians across a range of axes, including diagnostic accuracy, guideline entailment, management reasoning, and clinical communication skills.}
    \label{fig:study_design}
\end{figure}

\section{Evaluation}\label{sec:evaluation}

In this work, we sought to evaluate AMIE's capabilities in conversational AI for disease management, extending beyond diagnostic accuracy to encompass longitudinal reasoning and guideline-driven clinical decision-making. To this end, we adapted and extended the remote virtual OSCE paradigm introduced in \citet{tu2024towards} (\cref{sec:osce_study}).
To assess medication reasoning specifically, we developed RxQA, a multiple-choice question benchmark derived from drug formularies and validated by board-certified pharmacists (\cref{sec:rxqa_methods}).

\subsection{Multi-visit OSCE Study}
\label{sec:osce_study}

We employed a randomized, blinded design comparing AMIE to PCPs in the task of performing multi-visit consultations with trained patient actors via a synchronous text-chat interface (\cref{fig:study_design}). This approach allowed us to simulate the longitudinal nature of disease management and assess the ability to adapt plans and recommendations based on evolving patient information revealed in an incremental manner. Patient actors completed each multi-visit scenario twice---once with AMIE and once with a PCP---in a blinded and randomized order. Each scenario completion involved three text-chat conversations corresponding to visits 1, 2, and 3, with an interval of around 2 days (minimum 24 hours) between visits. The exact interval depended on the availability of patient actors and PCPs during a given week. For logistical reasons, the between-visit intervals enforced in the OSCE study were static and did not necessarily reflect the duration of time between visits specified by the narrative of the case scenario (e.g. follow-up after 2 weeks). Patient actors, PCPs, and AMIE were instructed to act based on the time intervals described in the scenarios. To simulate the patient-physician relationship and assess longitudinal reasoning, the same patient actor was used for all three visits of a given scenario and for both scenario completions (one with AMIE and one with a PCP), ensuring consistency in how scenarios were enacted. For PCP interactions, the same PCP was paired with the same patient actor for all three scenario visits.

\subsubsection{Scenarios}\label{sec:evaluation_scenarios}

Scenario packs were prepared by clinical providers in Canada and India and designed to describe the evolution of a patient's condition over three distinct visits. Each scenario specified the chief concern, symptoms, and medical history at the initial visit, as well as changes in these factors, treatment responses, test results, and imaging findings available at the beginning of subsequent follow-up visits. A total of 100 scenarios were used. The scenarios were equally distributed between providers in India and Canada, and across five medical specialties (20 scenarios each): cardiology, pulmonology, OB/GYN/urology, gastroenterology, and neurology. This balanced design ensured a comprehensive evaluation across diverse medical domains and clinical contexts. To assess the ability to handle complexities of real-world practice, scenarios were designed with varying difficulty levels, with many exhibiting information inconsistency (e.g., patient not sharing all information truthfully) and/or multi-morbidity (e.g., multiple co-presenting diseases requiring treatment trade-offs).

\subsubsection{Clinical Guidelines}
\label{sec:evaluation_guidelines}
\begin{table}[]
\centering
\captionsetup{justification=centering}
\caption{Size of the corpus of clinical guidelines used in the OSCE study.}
\vspace{0.5em}
\label{tab:guidelines_stats}
\begin{tabular}{l lll lll lll}
\toprule
           & \multicolumn{3}{c}{\textbf{NICE Guidance} (n=527)} & \multicolumn{3}{c}{\textbf{BMJ Best Practice} (n=100)} & \multicolumn{3}{c}{\textbf{All} (n=627)} \\
           & mean & max & total & mean & max & total & mean & max & total \\
\midrule
Characters & 62k & 168k &  33M & 138k & 205k & 13M & 74k & 205k & 46M \\
Tokens &  13k & 36k & 6M & 36k & 59k & 3M & 16k & 59k & 10M \\
\bottomrule 
\end{tabular}
\end{table}

A key aspect of the study design was the incorporation of clinical guidelines. Each scenario was designed to have a specific ground truth diagnosis, and was grounded in a pair of clinical practice guidelines, one from the UK National Institute for Health and Care Excellence (NICE) Guidance corpus\footnote{NICE Guidance: \url{https://www.nice.org.uk/guidance}} and one from the BMJ Best Practice corpus\footnote{BMJ Best Practice: \url{https://bestpractice.bmj.com}} respectively. NICE Guidance is a corpus of evidence-based recommendations for the health and social care sector, developed by independent committees, including professionals and lay members, and consulted on by stakeholders. BMJ Best Practice is a clinical decision support tool that supports healthcare professionals in making evidence-based decisions on diagnosis, treatment, and prevention. Where evidence is scarce or equivocal, expert opinion is provided in BMJ Best Practice. For each visit, the scenario specified a reference treatment plan consistent with these guidelines, outlining appropriate investigations and tests to perform during the visit, investigations to order after the visit, and action and treatment recommendations. This approach ensured that the consultations were grounded in established evidence-based practices, and allowed for testing the consistency of the systems involved with clinical guidelines.

The guideline corpus used in this study was a limited sample of the content covered by the NICE Guidance and BMJ Best Practice collections.
To simulate information resources and the lookup tasks faced by clinicians, both PCPs and AMIE had access to corpus of 627 clinical guidelines, comprising 527 NICE Guidance and 100 BMJ Best Practice documents. These included 50 NICE Guidance and 50 BMJ Best Practice guidelines that were directly relevant to the scenarios, as well as a sample of less relevant guidelines. Both PCPs and AMIE were instructed to reference this corpus during the study and ground their clinical decision-making in these guidelines as appropriate.
We report statistics about the corpus of guidelines in \cref{tab:guidelines_stats}.
Guideline content was used by AMIE at test time.

\subsubsection{Participants}

The study involved \numpcps board-certified PCPs and \numpatientactors validated patient actors, equally distributed across India and Canada. PCPs had a median of 9 years of post-residency experience (interquartile range of 5 to 12 years). To assess the quality of AMIE/PCP consultations and their responses to post-questionnaires, we recruited \numspecialists specialist physicians from India and North America, ensuring a range of clinical perspectives. These specialists were matched to evaluation tasks based on the medical specialty of the scenarios (e.g., cardiologists evaluated cardiology scenarios). Specialist physicians had a median of 5 years of post-residency experience (interquartile range of 4 to 8 years).

\subsubsection{Evaluation Measures}

For each scenario visit, data was collected from multiple sources, including patient actors, AMIE/PCPs, and specialist physicians.
Besides the consultation transcripts, data also included responses to post-questionnaires from AMIE and PCPs.
Post-questionnaires were completed by both AMIE (via offline generation) and PCPs (manually) after each visit.
Post-questionnaires asked for a differential diagnosis list (at least one and up to ten items), the set of clinical guidelines deemed most applicable (at least one and up to three from NICE Guidance and BMJ Best Practice each) and a management plan consisting of a list of investigations performed during the visit as well as investigations, treatments and follow-up actions recommended for after the visit.
To enable an assessment of MXEKF rubrics from a specialist physician perspective, the post-questionnaire also probed for aspects of management reasoning including whether and how AMIE/PCPs intentionally deviated from one of the selected clinical guidelines as well as comments about other acceptable management plans, competing priorities, the cost, effectiveness and side effects of the proposed plan, the prognosis with and without treatment of this patient, and recommendations for escalation and follow-up.
The evaluation framework considered both clinician-centered and patient-centered perspectives to provide a holistic assessment, in both cases with raters blinded to the source of the data (AMIE or PCP).

At the patient actor level, we assessed conversation quality and patient experience using rubrics from \citet{tu2024towards}, including the General Medical Council Patient Questionnaire (GMCPQ), evaluating aspects such as being polite and making the patient feel at ease; and the Practical Assessment of Clinical Examination Skills (PACES) rubric, assessing skills in eliciting information and managing patient concerns. However, we replaced the Patient-Centered Communication Best Practice (PCCBP) rubric from \cite{tu2024towards} with a new pilot evaluation rubric which we refer to as Management Reasoning Empirical Key Features (MXEKF). The MXEKF rubric (\cref{tab:mxekf_rubric_details}) was designed to specifically capture key aspects of management reasoning. MXEKF was inspired by prior work \cite{cook2023management} which identified twelve empirically-determined key features. Of these twelve features, we retained ten for MXEKF: \textit{(1) Contrasting and selection among multiple reasonable and defensible solutions}; \textit{(2) Prioritization of preferences, constraints, and values}; \textit{(3) Communication and shared decision making}; \textit{(4) Ongoing monitoring and adjustment of the management plan}; \textit{(5) Dynamic interplay among people, systems, settings, and competing priorities}; \textit{(6) Illness-specific knowledge}; \textit{(7) Clinician roles as patient teacher and salesperson}; \textit{(8) Clinician–patient relationship}; \textit{(9) Prognostication}; \textit{(10) Organization of the clinical encounter}. We opted not to include the other two features identified in \cite{cook2023management}: \textit{Process Knowledge} (assessing understanding of systems of care) because it would be specific to particular clinical environments, and \textit{Management Scripts} (understanding of conceptual knowledge structures and clinician tasks) because management scripts were infeasible to collect from PCPs as part of the OSCE study. The MXEKF rating questions are detailed in~\cref{tab:mxekf_rubric_details}. Patient actors completed the full GMCPQ rubric and subsets of the PACES and MXEKF rubrics after each of the three scenario visits.

At the specialist physician level, evaluations focused on consultation quality, guideline adherence, and the appropriateness of diagnostic and management decisions. Specialist physicians assessed AMIE and PCPs based on their consultation transcripts and responses to the post-questionnaires using various rubrics. In addition to the full MXEKF and PACES rubrics, these included an expanded version of the `Diagnosis \& Management' rubric from \citet{tu2024towards}. The additional rating questions in this expanded rubric are detailed in~\cref{tab:management_extended_rubric_details}. Beyond accuracy and completeness of the differential diagnosis, and the appropriateness of escalation, investigations, and treatments, the expanded rubric incorporated detailed evaluation axes related to guideline entailment and implementation, the preciseness of recommended investigations and treatment, and the ability to remember relevant information from prior visits.
Specialist physicians rated all rubric items for each visit separately (and considering past visits). During the evaluation task, information for all three visits per scenario were presented to specialist physicians at once to enable a holistic assessment under full information.
The evaluation interface for specialist physicians is shown in \cref{appendix:eval_ui}.

\subsection{RxQA Medication Reasoning Benchmark}
\label{sec:rxqa_methods}

To assess AMIE's medication reasoning further, we developed a multiple choice question benchmark, RxQA, which consists of 600 questions derived from national drug formularies across two jurisdictions: OpenFDA and the British National Formulary (BNF).\footnote{OpenFDA: \url{https://open.fda.gov/}; BNF: \url{https://bnf.nice.org.uk/}} We used Gemini 1.5 Flash \cite{geminiteam2024gemini15unlockingmultimodal}, with access to these medication labels, to draft and filter these questions, which were further refined by board-certified pharmacists in each jurisdiction (\cref{rxqa_prompts}). The dataset was built as follows:
\begin{enumerate}
    \item \textbf{Medication Labels:} We ingested the OpenFDA and BNF drug formularies, formatting the medication labels in a clean format and sampling a subset of labels from each source.
    \item \textbf{Question Generation:} From each label, we generated five short questions, each with four answer choices, using the prompt shown in \cref{easy_question_generation_prompt}, and three longer questions using the process shown in \cref{hard_question_generation_prompt}.
    \item \textbf{Question Filtering:} We validated question quality by asking Gemini to confirm each question was adequately specified and the listed answer was unambiguously correct among the four choices.
    \item \textbf{Question Selection:} We randomly selected 100 ``short questions'' and 200 ``long questions'' from each jurisdiction for which Gemini required the corresponding medication context to answer correctly.
    \item \textbf{Pharmacist Revision:} Finally, we recruited 4 board-certified pharmacists in each jurisdiction to revise the question wording, answer options, and the correct answer. The pharmacists rated the difficulty of each question (Trivial, Easy, Medium, Hard, Impossible). For subgroup analysis purposes, Trivial and Easy were mapped to lower difficulty, while Medium, Hard, and Impossible mapped to higher difficulty.
\end{enumerate}

An example question, derived from OpenFDA, is shown in \cref{fig:rxqa_example_question}. We assess AMIE's performance on RxQA by ingesting the OpenFDA and BNF drug formularies with the Mx Agent, and prompting the system with the question in a zero-shot manner. As a baseline of human test-taking performance, we also asked a group of 3 primary care physicians in each jurisdiction to answer each question, first without the reference medication context and then with access to this context.

\subsection{Statistical Analysis}

The quality of management plans for each of three visits was measured as the proportion of cases with favorable ratings from specialist physicians.
Of 15 evaluation axes tested, 9 were based on Yes/No ratings scales. The remaining ones were binarized using the top-2 options on their respective scales: `Overall Appropriate' (5-point scale), `Selected Applicable Guidelines' (5-point scale), `Aligned with Guidelines' (5-point scale), `References Guidelines' (4-point scale).
For each evaluation axis, cases with `N/A' ratings on either study arm were excluded for each visit.
The McNemar test was used for each evaluation axis and visit and p-values were adjusted using the false discovery rate (FDR) correction.

Relative performance of PCPs and AMIE on each of the 10 MXEKF evaluation axes was measured in terms of preferences expressed by specialist physicians and patient actors respectively.
Preferences were derived from independent ratings (on a 5-point scale ranging from `Poor' to `Excellent') for each of the 10 axes and each of three visits per scenario.
For 3 of the 10 MXEKF evaluation axes (Contrasting and Selection, Illness-Specific Knowledge, Prognostication), ratings were collected only from specialist physicians, but not patient actors.
In total, we computed preference rates for 51 unique combinations of MXEKF axis, scenario visit and rater perspective (3 visits $\times$ 10 MXEKF axes for specialist physicians $+$ 3 visits $\times$ 7 MXEKF axes for patient actors).
Summary metrics for preference rates refer to the spread of preferences across these 51 sets of preference rates.

For comparisons of RxQA medication reasoning accuracy, the McNemar test was used with question-level pairing and p-values were adjusted using FDR correction.

\begin{figure}[ht!]
    \centering
    \includegraphics[width=\textwidth,height=\textheight,keepaspectratio]{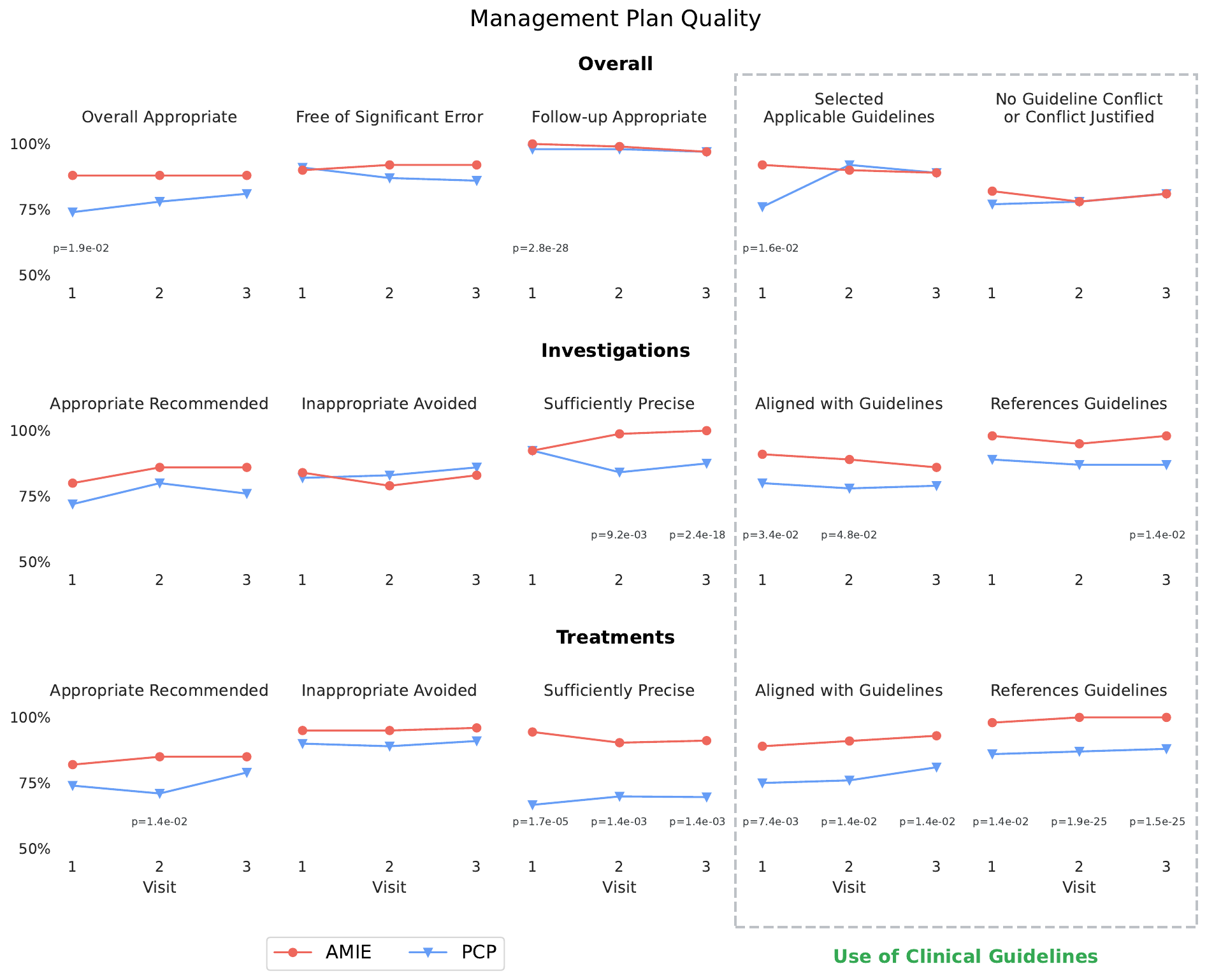}
    \vspace{0.5cm}
    \caption{\textbf{Management plan quality.} The quality of management plans for each of three visits per scenario, measured as the proportion of cases where AMIE or PCPs received favorable ratings from specialist physicians. We present overall quality criteria alongside criteria specific to investigations and treatments respectively. For each category, we include two quality criteria address the use of clinical guidelines (right). Of 15 evaluation axes tested, 9 were based on Yes/No ratings scales. The remaining ones were binarized using the top-2 options on their respective scales: `Overall Appropriate' (5-point scale), `Selected Applicable Guidelines' (5-point scale), `Aligned with Guidelines' (5-point scale), `References Guidelines' (4-point scale). For each evaluation axis, cases with `N/A' ratings on either study arm were excluded for each visit.
    The sample size is N=100 scenarios for all evaluation axes and visits, except `Sufficiently Precise' (Investigations: N=92, N=82, N=72 for three visits; Treatments: N=72, N=83, N=79 for three visits). P-values from McNemar test are shown for all comparisons with $p<0.05$ after false discovery rate (FDR) correction.}
    \label{fig:management_plan_quality_by_visit}
\end{figure}

\section{Results}
\label{sec:results}

\subsection{Management Plan Quality}

The evaluation axes shown in Figure \ref{fig:management_plan_quality_by_visit} pertain to three categories, with five evaluation axes per category: the \textit{overall} quality of the management plan, the quality of \textit{investigation} recommendations, and the quality of \textit{treatment} recommendations.
Each of the three categories includes two evaluation axes specifically related to the \textit{use of clinical guidelines} (right-hand side).

\subsubsection{AMIE's management plans were non-inferior to those from PCPs.}

Across all 15 evaluation axes and three visits, AMIE's management plans scored at least as well as those from PCPs.
For the first visit, AMIE scored significantly higher than PCPs for overall appropriateness of the plan (88\% vs. 74\%, p=0.019) and providing appropriate follow-up recommendations (100\% vs. 98\%, p<0.001).
However, performance for these axes assimilated during follow-up visits and differences were not significant.
AMIE also scored higher for recommending appropriate treatments, but the difference was only significant for the first follow-up visit (85\% vs. 71\%, p=0.014).
PCPs scored marginally higher than AMIE on some axes, e.g., avoiding inappropriate investigations during follow-up visits (79\% vs. 83\%, p=0.716 in visit 2; 83\% vs. 86\%, p=0.862 in visit 3), but differences were not statistically significant.

\subsubsection{AMIE was more precise in recommending investigations and treatments.}

One key strength of AMIE was the level of preciseness in recommending investigations and treatments at the end of each visit.
For treatments, AMIE received higher preciseness scores than PCPs consistently during all three visits (94\% vs. 67\%, p<0.001 in visit 1; 90\% vs. 70\%, p=0.001 in visit 2; 91\% vs. 70\%, p=0.001 in visit 3).
For investigations, preciseness scores were on par at 91\% for the initial visit.
However, for follow-up visits, AMIE's preciseness scores improved whereas PCP scores declined, leading to a statistically significant gap (99\% vs. 84\%, p=0.009 in visit 2; 100\% vs. 88\%, p<0.001 in visit 3).

\begin{figure}[ht!]
    \centering
    \includegraphics[width=\textwidth,height=\textheight,keepaspectratio]{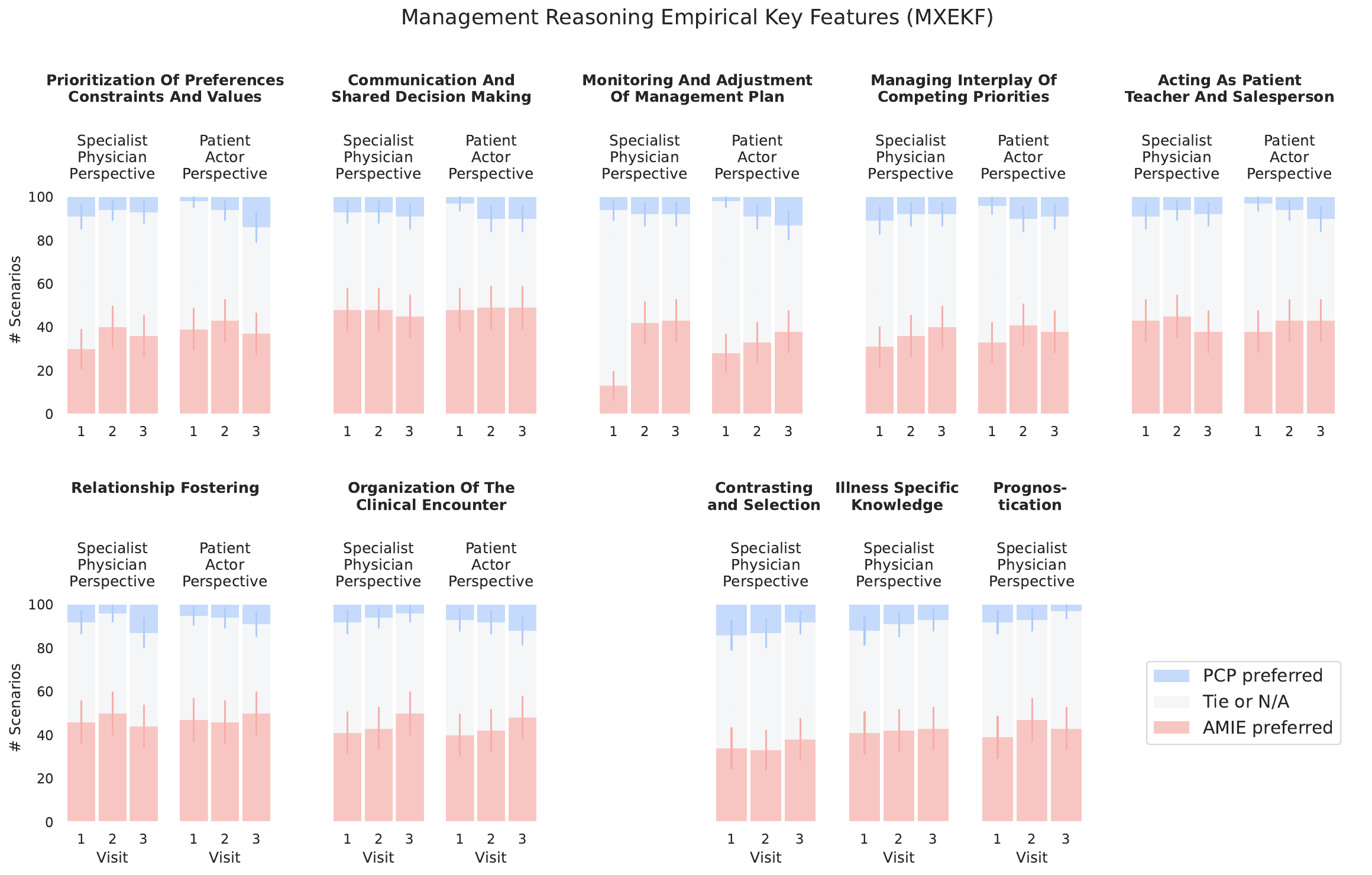}
    \vspace{0.5cm}
    \caption{\textbf{Management Reasoning Empirical Key Features (MXEKF).} Relative performance of PCPs and AMIE on each of the 10 MXEKF evaluation axes in terms of preferences expressed by specialist physicians and patient actors respectively. Preferences were derived from independent ratings (on a 5-point scale ranging from `Poor' to `Excellent') for each of three visits per scenario. For 3 of the 10 MXEKF evaluation axes (Contrasting and Selection, Illness-Specific Knowledge, Prognostication), ratings were collected only from specialist physicians. Cases with identical ratings and those including one or more N/A ratings were grouped together as `Tie or N/A'. Error bars represent 95\% confidence intervals for binomial proportions.}
    \label{fig:mxekf_preferences_by_visit_both_perspectives}
\end{figure}

\subsubsection{AMIE's management plans were better aligned with and grounded in clinical guidelines.}

As pertaining to the use of clinical guidelines, AMIE scored significantly higher than PCPs for selecting applicable guidelines on the first visit (92\% vs. 76\%, p=0.016). However, PCPs made up for this gap in follow-up visits where no differences could be detected.
For all three visits, AMIE received significantly higher scores for recommending treatments that were \textit{aligned} with the guidelines (89\% vs. 75\%, p=0.007 in visit 1; 91\% vs. 76\%, p=0.014 in visit 2; 93\% vs. 81\%, p=0.014 in visit 3) and supported by \textit{explicit references} to a guideline (98\% vs. 86\%, p=0.014 in visit 1; 100\% vs. 87\%, p<0.001 in visit 2; 100\% vs. 88\%, p<0.001 in visit 3).
Similar, though less consistent trends were observed for recommending investigations that were \textit{aligned} with the guidelines (91\% vs. 80\%, p=0.034 in visit 1; 89\% vs. 78\%, p=0.048 in visit 2) and supported by \textit{explicit references} to a guideline (98\% vs. 87\%, p=0.014 in visit 3).

\subsection{Management Reasoning Empirical Key Features (MXEKF)}

Relative performance of PCPs and AMIE on each of the 10 MXEKF evaluation axes was measured in terms of preferences expressed by specialist physicians and patient actors respectively. Figure \ref{fig:mxekf_preferences_by_visit_both_perspectives} visualizes preference rates for a total of 51 unique combinations of MXEKF axis, scenario visit and rater perspective (3 visits $\times$ 10 MXEKF axes for specialist physicians $+$ 3 visits $\times$ 7 MXEKF axes for patient actors).

\begin{figure}[ht!]
    \centering
    \includegraphics[width=0.9\textwidth,height=\textheight,keepaspectratio]{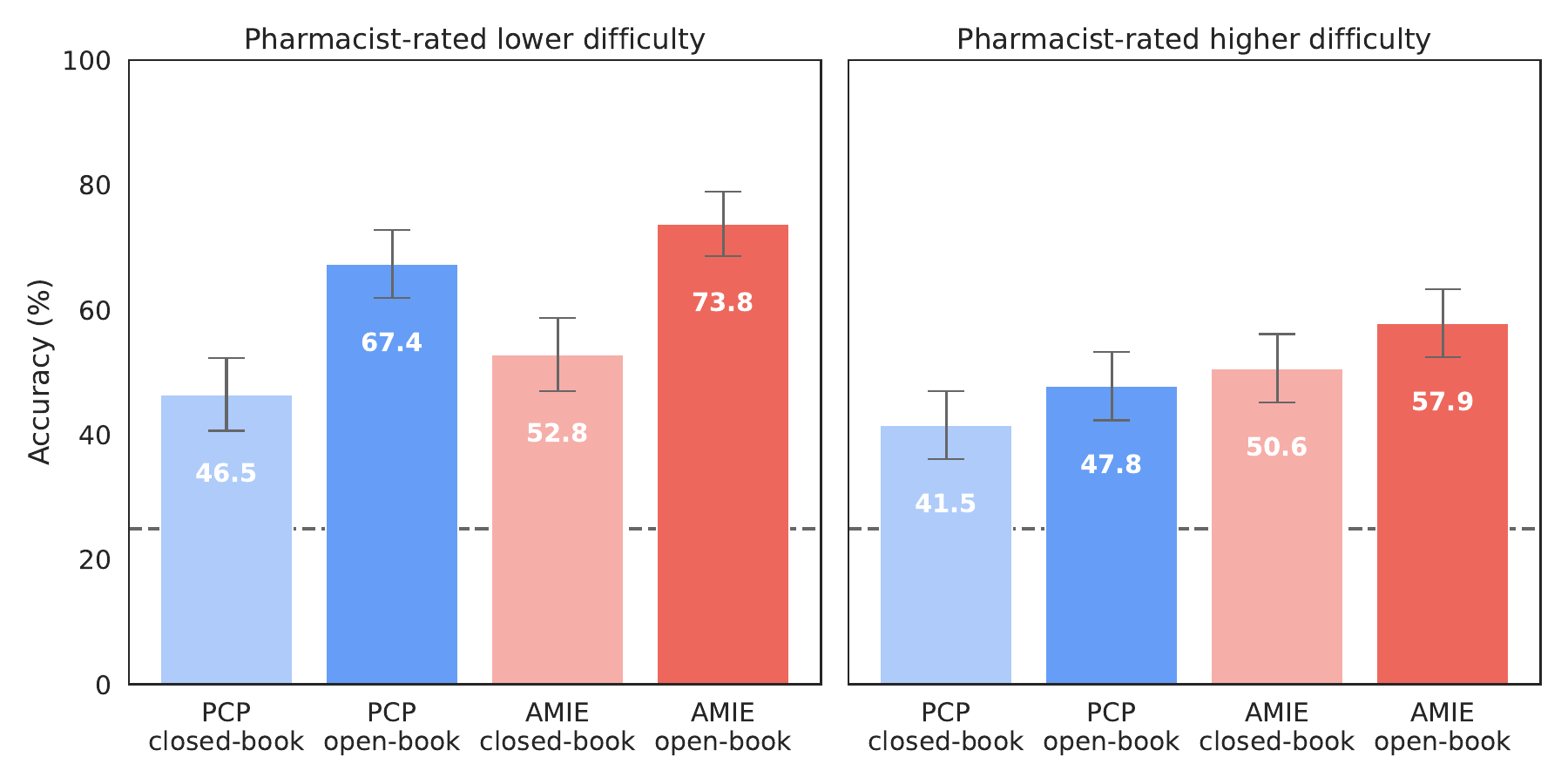}
    \vspace{0.5cm}
    \caption{\textbf{RxQA Medication Reasoning Accuracy.} Question-answering accuracy of PCPs and AMIE on the RxQA medication reasoning benchmark. Results are presented separately for lower-difficulty (left) and higher-difficulty (right) questions as rated by a set of board-certified pharmacists. Here, we consider questions which pharmacists rated as ``Trivial'' or ``Easy'' to be lower-difficulty questions. We compare a ``closed-book'' setting in which neither PCPs nor AMIE have access to external knowledge resources, to an ``open-book'' setting where PCPs are given the relevant medication label for each question, and AMIE is permitted to retrieve relevant information from drug formularies. Error bars represent 95\% confidence intervals for binomial proportions. The dashed horizontal line indicates accuracy achievable through random guessing.}
    \label{fig:rxqa_accuracy}
\end{figure}

\subsubsection{AMIE was preferred more often than PCPs by patient actors and specialist physicians.}

In roughly half of cases, no clear preference could be determined between AMIE and PCPs, resulting in a median tie rate of 50\% (95\% CI, 41\%-68\%; N=51).
However, when a preference was expressed, AMIE, with a median win rate of 42\% (95\% CI, 29\%-50\%; N=51), was preferred significantly more often than PCPs who had a median win rate of only 8\% (95\% CI, 2\%-14\%; N=51).
This trend was consistent across specialist physician and patient actor perspectives. See \cref{sec:distribution_of_patient_actor_ratings,sec:distribution_of_specialist_physician_ratings} for the full distributions of patient actor and specialist physician ratings respectively.

\subsubsection{MXEKF scores evolve across visits.}

For certain MXEKF axes, especially those pertaining to the temporal aspects of management reasoning, preference rates exhibited distinctive patterns across the three visits per scenario.
For example, preferences for `Monitoring and Adjustment of Management Plan' became more pronounced over the course of multiple visits, with tie rates shrinking, and preference rates increasing for both PCPs and AMIE and both rater perspectives.
For `Communication and Shared Decision Making', preference rates for PCPs increased across visits while AMIE's scores remained consistent at a high level.
For axes assessing the `Organization of the Clinical Encounter' and demonstration of `Illness-specific Knowledge', specialist physician preferences increased in favor of AMIE while decreasing for PCPs across three visits.

\subsection{Medication Reasoning Accuracy}

We present medication reasoning accuracy on the RxQA benchmark separately for lower-difficulty (N=282) and higher-difficulty (N=318) questions in \cref{fig:rxqa_accuracy}.
We compare a ``closed-book'' setting in which neither PCPs nor AMIE had access to external knowledge resources, to an ``open-book'' setting where PCPs were given the relevant medication labels for each question, and AMIE was permitted to retrieve relevant information from the OpenFDA and BNF drug formularies.
Table \ref{tab:rxqa_accuracy_stats} provides detailed results for all comparisons.

\subsubsection{RxQA is challenging for both PCPs and AMIE.}

We observed that RxQA was a challenging test for both AMIE and PCPs, leaving ample headroom for future improvements.
While test accuracy differed significantly between closed-book and open-book settings as well as difficulty levels, peak performance was below 75\% for both PCPs (67.4\%) and AMIE (73.8\%) even in the lower-difficulty open-book setting.

\subsubsection{AMIE was more accurate than PCPs in higher-difficulty questions.}

For the subset of questions rated by pharmacists as higher difficulty, AMIE was significantly more accurate than PCPs in both the closed-book setting (50.6\% vs. 41.5\%, p=0.013) and the open-book setting (57.9\% vs. 47.8\%, p<0.001).
No significant difference was detected for questions of pharmacist-rated lower difficulty, neither for the closed-book setting (52.8\% vs. 46.5\%, p=0.147) nor for the open-book setting (73.8\% vs. 67.4\%, p=0.071).

\subsubsection{Access to external knowledge resources helps especially in lower-difficulty questions.}

Both PCPs and AMIE benefited significantly from the ability to access external knowledge resources.
For lower-difficulty questions, both PCPs (46.6\% vs. 67.4\%, p<0.001) and AMIE (52.8\% vs. 73.8\%, p<0.001) significantly increased their accuracy (by more than 20\%) when transitioning from closed-book to open-book setting.
We observed the same effect for higher-difficulty questions.
However, differences were less pronounced, yet statistically significant for both PCPs (41.5\% vs. 47.8\%, p=0.045) and AMIE (50.6\% vs. 57.9\%, p=0.010).

\section{Related Work}
\label{sec:related-work}

\subsection{Management Reasoning}

Early work in management reasoning focused on the adaptation of decision theory to medicine in easily categorized decisions with clear ``right'' answers, such as laparotomy in acute abdomen \cite{De_Dombal1974-gd} or antibiotic selection for already-identified microorganisms \cite{shortliffe1987computer}. More recent approaches have been  grounded in cognitive psychology, in particular how human clinicians process and store clinical information \cite{Elstein1988CognitivePI}. Clinicians group both diagnostic and management information into semantic structures known as ``scripts'' \cite{Bordage1991-xn}. The nuance and accuracy of scripts is highly influenced by experience, but also subject to considerable contextual factors as well as cognitive biases, heuristic short cuts that can lead to diagnostic and management errors~\cite{Croskerry2009-or}. 

\subsection{Management Reasoning by AI Systems}

Early management reasoning systems used Bayesian inference and rules-based systems to model management decisions in limited domains \cite{Adams1986-xu}. LLMs show human-like abilities in many specific management tasks, such as triaging emergency room patients for admission \cite{Williams2024-nz}, extracting oncological information from radiology reports \cite{Chen2024-pf} or navigating general clinical encounters with uncertainty \cite{Goh2025-wa}. They perform less well in more generalized reasoning tasks, erring on the side of caution in high-stakes settings \cite{Williams2024-dc} or failing to select the appropriate medical calculators for clinical prediction tasks \cite{Khandekar2024-ih}. Computational technologies such as retrieval-augmented generation \cite{lewis2021retrievalaugmentedgenerationknowledgeintensivenlp} (RAG) allow for models to take advantage of clinical guidelines and other expert-curated materials; RAG-based systems outperform base models on a variety of clinical tasks \cite{Zakka2024-ss} and hold considerable promise for management reasoning. Data from clinical trials is lacking for LLM clinical decision support systems; a number of trials have been registered and are recruiting patients, but a several factors, including privacy rules, have slowed study \cite{Omar2024-yc}.

\subsection{Conversational AI and Goal-oriented Dialogue}

The current ability of LLMs to prospectively collect data is hotly debated; automated evaluations using AI agents show that the diagnostic accuracy of general purpose LLMs falls considerably when actively collecting information compared to being presented with vignettes or clinical documentation \cite{Johri2025-kr}. However, conversational models have already been clinically deployed \cite{Mukherjee2024-yb}, and data from a virtual primary care clinic shows that conversational AI approaches human diagnostic accuracy for a variety of common primary care concerns, though with considerable heterogeneity by diagnosis \cite{Zeltzer2023-iu}. Real-world integration of an AI agent into patient chats show high levels of accuracy and safety, with improved patient satisfaction compared to the previous human baseline \cite{Lizee2024-uy}. While these systems integrate hybrid clinician/AI teams in conversations with patients, AMIE has shown superior diagnostic accuracy and patient-centric behavior in text-based conversation with patient actors compared to human PCPs across a broad array of diagnoses \cite{tu2024towards}. 

\subsection{AI for Medical Consultations}

Providing LLMs to generalist physicians improved management in difficult cases with no right answers, largely by prompting physicians to better consider patient-specific factors \cite{Goh2025-wa}. When instructed to solve complex subspecialist cardiology cases with AMIE, the management quality of general cardiologists increased in 64\% of cases, decreasing in only 3\% of cases~ \cite{o2024towards}. In particular, AMIE served as an initial, highly sensitive diagnostic screen, with generalist knowledge serving as a highly specific confirmatory test to increase accuracy. 

\subsection{Evaluation of Dialogue for Disease Management}

The most common evaluation method for clinical dialogue is the Objective Structured Clinical Examination (OSCE)~\cite{Harden1975-wn}. OSCEs are simulated patient encounters, in which a  student performs a diagnostic and management interview with a trained patient actor. OSCEs cover a variety of encounter types, including counseling, end-of-life conversations, and trauma-informed care, as well as more traditional diagnostic and management encounters. Due to their similarity to actual clinical encounters and reliability, OSCEs have become commonplace in physician licensing assessments across the world, including the USMLE in the United States and PACES in the UK \cite{Gentles2003-ip}. The dynamic nature of clinical conversations makes ``ground truth'' evaluation difficult, so OSCEs are graded both via case-specific rubrics and global assessments \cite{Ogunyemi2017-bx,Malau-Aduli2012-nj}.

\subsection{Multi-Agent Systems and Action Planning}

Multi-agent systems can be separated into two groups: simulated environments in which autonomous AI agents interact \cite{park2023generativeagentsinteractivesimulacra, li2025agenthospitalsimulacrumhospital, vezhnevets2023generativeagentbasedmodelingactions}, and cooperative multi-specialist systems \cite{shen2023hugginggptsolvingaitasks, mukherjee2024polaris, gottweis2025aicoscientist, christakopoulou2024agents}. \citet{mukherjee2024polaris} introduced Polaris, a constellation of specialists (for checklists, medications, labs, vitals, etc.) supporting a single user-facing agent with promising performance in medication safety, clinical conversations, and bedside manner when compared with nurses. \citet{gottweis2025aicoscientist} studied a cohort of specialized agents capable of reasoning and interacting with tools to explore scientific problems. \citet{christakopoulou2024agents} proposed a system distributing work between a fast user-facing system 1 (Talker) agent, and a slower system 2 (Reasoner) in the context of sleep coaching. The Reasoner plans the conversation ahead by autonomously interacting with tools and memory. Planning actions using LLMs has been explored in the context of embodied agents \cite{wang2024describeexplainplanselect,ahn2022icanisay,wang2023voyageropenendedembodiedagent}. AMIE's Mx Agent is an example of an agent planning actions (e.g., investigations or treatments) to inform the care journey and which are delivered in a conversational manner through the Dialogue Agent.

\subsection{Reasoning, Test-Time Scaling and Knowledge Exploration}

Foundation models and reasoning techniques, particularly extensions of chain-of-thought, self-consistency and refinement~\cite{wei2022chain, kojima2022large, wang2022self, madaan2023self}, have long been leveraged in challenging medical benchmarks \cite{lievin2022can, singhal2023large, singhal2023towards, nori2023can} like the MedQA-USMLE~\cite{jin2021disease}. Building upon these foundational approaches, agentic frameworks that interleave reasoning and retrieval~\cite{yao2023react, shinn2023reflexionlanguageagentsverbal, gao2024retrievalaugmentedgenerationlargelanguage, singh2025agenticretrievalaugmentedgenerationsurvey} have driven further advancements in science and medicine. Med-Gemini and AI co-scientist~\cite{saab2024capabilities, gottweis2025aicoscientist} demonstrated further improvements by scaling test-time computation~\cite{snell2024scaling, muennighoff2025s1}, enabling a broader exploration of the solution space. Operating under real-time dialogue constraints, AMIE's Mx Agent simultaneously explores multiple patient care strategies and leverages its long-context capabilities to refine these intermediate solutions using extensive authoritative knowledge. We investigated the use of structural constraints to enhance control over the reasoning process, a technique also explored in \cite{zhou2024selfdiscoverlargelanguagemodels, tam2024letspeakfreelystudy}. While structural constraints are valuable for tasks involving complex outputs and time constraints, such as real-time patient management, the most powerful reasoning methods may be those that can automatically, predictably, and adaptively scale their cognitive processes to the demands of the task. Recent advancements in emerging ``thinking'' models such as Gemini 2.0 Flash Thinking~\cite{google-flash-thinking}, OpenAI-o1\cite{openai-o1}, and DeepSeek-R1~\cite{deepseekai2025-R1} are exploring this frontier.

\section{Discussion}
\label{sec:discussion}

In this study, we showed that AMIE performed consistently well across a variety of management reasoning challenges, conducting clinical conversations that spanned multiple visits per patient and producing management plans that were non-inferior to board-certified PCPs.

AMIE's non-inferior performance compared to PCPs was maintained throughout longitudinal clinical encounters. For the initial consultation in particular, AMIE outperformed PCPs in the overall appropriateness of the plan and providing appropriate follow up recommendations, though the performance gap closed with PCPs over the subsequent encounters. There were no domains in which PCPs outperformed AMIE. AMIE's performance in longitudinal care settings, including successfully interpreting additional diagnostic information such as laboratory results and information from other medical consultants, is an important test of a potential real-world use case for conversational AI. Health systems worldwide are experiencing increased care fragmentation which is associated with worsened morbidity for patients with chronic illnesses. In the real world, it is often more likely that a patient may have seen different clinicians over sequential visits~\cite{Joo2023-df}. AI systems have not previously been specifically evaluated for this crucial unmet need. Our results suggest that AMIE and other conversational management agents might one day be able to serve as a point of continuity in otherwise fragmented health systems, either independently or working with clinicians. If such systems can be developed and tested with care and robust clinical evidence, they have the potential to address growing unmet clinical needs caused by global shortages and inequalities of supply of PCPs, exacerbated by concerns of physician burnout and increasingly complex populations under management~\cite{Lawson2023-jd}.

AMIE's management plans had some important differences from those of human clinicians; for example, AMIE was considerably more precise in both recommending treatments and investigations, a trend that held over all three clinic visits. The improved preciseness---giving specific and clear instructions for medications, follow up plans, and contingencies---manifests as a change from generalized recommendations to actionable clinical directives. Previous versions of AMIE~\cite{tu2024towards} might have offered broad suggestions, such as ``prescribe an antibiotic'' for a bacterial infection or ``recommend a follow-up appointment'' for a patient requiring further monitoring. While directionally correct, such recommendations have limited practicality in actual clinical care situations. In contrast, the enhanced variant of AMIE proposed in this work generates plans with a level of granularity that directly supports clinical implementation, such as naming the precise antibiotic, its dose, the duration of treatment and administration route, and any monitoring or follow-up actions that are required. Beyond clinical acceptability, preciseness has important implications for care engagement, especially given the increasing movement for patients to have access to their own medical notes. Precise, clear language is preferred by patients \cite{Vanka2025-ur} and might drive increased completion of follow up visits and ordered diagnostics and tests 
\cite{Bell2024-ll}. 

AMIE also selected the appropriate guidelines better than PCPs for the first visit, and its management decisions were better aligned with guidelines for all three visits. Management reasoning encompasses far more than entailment in guidelines, which often do not apply to individual practice settings, and experienced clinicians more often deviate from guidelines than novices~\cite{Mercuri2015-xw, Woolf1999-so}. However, implementation of clinical practice guidelines has been associated with higher quality of care \cite{Lugtenberg2009-jg} and for increasingly complex chronic illnesses, can reduce patient morbidity and mortality \cite{McCullough2021-wr}. AMIE's guideline selection and alignment holds promise for increasing the feasibility of quality initiatives that might seek to implement guideline-directed medical management of chronic medical conditions such as congestive heart failure and diabetes mellitus, or to assist in the study of unwarranted variation from such recommendations. AMIE's use of clinical guidelines  also allows for considerable flexibility and customization as these guidelines can be easily updated; a clinic in a low- and medium-income country, for example, could easily switch from NICE Guidance or BMJ Best Practice guidelines to World Health Organization guidelines more relevant to their care setting. Similarly, drug shortages have continued to be a global problem affecting health systems \cite{Shukar2021-ej} and can present challenges to the correct implementation of practice guidelines; health systems utilizing AMIE could just as easily update their guidelines as they provide practice updates to clinicians. 

In this study, we also developed the Management Reasoning Empirical Key Features (MXEKF) rubric, derived from psychological research on outpatient clinicians \cite{Cook2018-xs}. This pilot evaluation rubric addresses one of the major challenges to measuring management reasoning quality---that decisions are subject to considerable amounts of context specificity, and that there is often no single answer to the best management plan. LLMs have generally fared poorer on management tasks compared with diagnostic tasks than humans. However, on blinded ratings of MXEKF by specialist physicians, AMIE was preferred to PCPs significantly more often than PCPs were preferred, although both were tied in about half of the cases. These results require further exploration in follow up studies, but suggest AMIE's potential for exhibiting PCP-like quality in management reasoning.

AMIE's abilities reflect the growing capabilities of LLMs for clinical conversation and reasoning. Its capabilities are intrinsically linked to advances across various domains, including reasoning \cite{google-flash-thinking, openai-o1, deepseekai2025-R1} and long-context processing \cite{geminiteam2024gemini15unlockingmultimodal, saab2024capabilities, lee2024can}. Because of this, we suspect that current limitations---in particular, confabulations, which offer considerable risks in clinical medicine---may be mitigated through the steady improvement of the state-of-the-art LLMs underpinning the AMIE system. AMIE's multi-agent system, as well as its ability to cite guidelines, appears to have had a significant impact: confabulation rates are not only comparable between PCPs and AMIE but also considerably lower than those reported in other studies  \cite{singhal2023large, 10.1001/jamanetworkopen.2023.47075}. The rapid progress of state-of-the-art LLMs holds promise for factual summarization. While our backbone model, Gemini 1.5 Flash, exhibits a hallucination rate above 5\% on the HHEM leaderboard~\cite{vectara_hallucination_leaderboard}, the updated Gemini 2.0 Flash reduced this rate to 0.7\%.

We also introduced a novel benchmark for medication reasoning, RxQA. The saturation of traditional medical reasoning benchmarks has added urgency to the development of new and more discriminant benchmarks \cite{Brodeur2024-nv, Raji2025-bf}.  It also underscores a new reality---an era of novel capabilities of AI systems with human-like performance in domains that were once thought out of reach of machines. RxQA benchmarks the knowledge of and reasoning about the use of medications. Derived from the national drug formularies of two countries, it is not intended to be fully comprehensive, but rather to be updated over time with local data. RxQA was a challenging benchmark for both AMIE and human PCPs, with equivalent performance on the less difficult questions. However, AMIE outperformed humans on the more difficult questions compared to PCPs, both open- and closed-book.

Despite these promising results, our study has multiple limitations. While patient actors are a gold standard in medical education for the assessment of trainee clinicians, they are not representative of clinical care. Clinical scenarios are constructed---meaning that they have definitive answers---and constrained for reliability in scoring. Because of this, cases lack much of the chart review to determine clinical history that characterizes real patient care. The case mix is also not indicative or representative of a real clinical practice setting. Our cases utilized guidelines from the UK, while our PCPs were from North America and India. While PCPs were given relevant guidelines to review for each case as a method of mitigating these differences, this is not fully consistent with clinical practice, in which clinicians are unconstrained by guidelines and use their personal experience or point-of-care reference tools to aid in management decisions.

While our cases unfolded over the course of weeks or months in the narrative specified in scenarios, the actual time between visits in our study was 1-2 days. This likely increased human performance on these cases, as it is not a realistic test of human memory. The interval history was also considerably constrained to the patient's main complaint, far different than the open-ended paths of real patient care.

While utilizing patient actors in an OSCE format mirrors standards for evaluating physicians' consultations, our results require further contextualization. Patient actors communicated with both AMIE and PCPs via a text-based chat interface. Despite the increased popularity of text-based patient portals for communication with PCPs \cite{Carini2021-mg}, telehealth visits are generally conducted either via audio or video chat and are by nature multimodal. Telehealth has grown in both popularity and importance since the COVID-19 pandemic \cite{Shaver2022-yb}, spurring the development of telehealth OSCEs for health professions students~\cite{Sartori2020-ra}. The use of a text-only interface for AMIE was selected because of the convenience of this modality in communicating with a chatbot, and relative uniformity of the interaction compared to complexities of intonation, body language, and other cues in a multimodal audio/visual setting. It would further be significantly harder to blind both the patient actors and specialist raters to the source of the conversation in a multimodal setting. 

The user interface configured for AMIE in this study has considerable differences to actual patient care. As above, telehealth is generally conducted via audio or video chat, not via a text interface. Orders are not entered via free text but via an electronic provider order entry system. This is not only an important agentic task, but also constrains order entry in a manner that likely would improve AMIE's performance because these systems often contain order sets and only allow medication from acceptable drug formularies, including pre-populated doses and clinical decision support. Practice alerts, as well as oversight from pharmacists, in health systems also serve to improve accuracy, which is absent in our study.

While we have attempted to develop and validate a tool for global measures of management reasoning that has construct validity, more work is necessary to demonstrate the reliability of MXEKF in the real world, as well as further describing its psychometric characteristics. This should be seen as a first step in the measurement of management reasoning.

The RxQA benchmark also has several limitations. Selecting questions for which Gemini initially required medication labels to select the correct answer likely skewed the questions to be relatively harder and not necessarily representative of typical practice. While each question was revised and validated by a board-certified pharmacist, there could be significant inter-pharmacist variability. Furthermore, while we benchmark against PCPs, in the real world, medication decisions are often made with guidance and oversight from pharmacists. We do not suggest that these examples of human performance indicate any measure of real-world competence, but rather intend for them to serve as an experimental control group, or baseline assessment to help contextualize comparisons of various AI systems in medication knowledge retrieval tasks. 

\section{Conclusion}
\label{sec:conclusion}

In this work, we demonstrated AMIE's ability to not only take histories and form differential diagnoses, but make nuanced management decisions that meet or exceed those of PCPs, including in cases that occur over several visits. Our findings do not suggest that AMIE is ready for clinical care. Our study fundamentally explores the art of the possible, and many additional steps---including prospective feasibility studies on patients with appropriate ethical and safety oversight---are necessary to ensure that AMIE can function as part of a health care team. Nonetheless, this study represents a milestone towards the goal of safe, equitable, and ethical agentic health AI systems that can scale quality healthcare in increasingly fragmented health systems.

\subsubsection*{Acknowledgments}
This project was an extensive collaboration between many teams at Google Research and Google DeepMind.
We thank Abhijit Guha Roy, Yun Liu, Yishay Mansour, Rachelle Sico and Joelle Wilson for their comprehensive review and detailed feedback on the manuscript.
We also thank John Guilyard, Brian Gabriel and Jenn Sturgeon for contributions to the narratives and visuals.
We are grateful to Rachel Gruner, Bakul Patel, Jinhyuk Lee, Jon Shlens, Sally Goldman, Ajay Joshi, Yuri Vasilevski, Sean Li and Sherol Chen for their valuable insights, technical support and feedback during our research.
We also thank SiWai Man, Brett Hatfield and Gordon Turner for supporting the OSCE study overall, GoodLabs Studio Inc, Intel Medical Inc and Chris Smith for their partnership in conducting the OSCE study in North America, and JSS Academy of Higher Education and Research and Vikram Patil for their partnership in conducting the OSCE study in India.
We are grateful to our partners at the UK National Institute for Health and Care Excellence, and BMJ Best Practice for their partnership on clinical guideline content.
We thank our partners at the Royal Pharmaceutical Society for supporting the creation of the RxQA benchmark through content from the British National Formulary.
We also thank Shruti Garg, Fahd Malik and Edgar Camelo for facilitating these partnerships.
Finally, we are grateful to David Barrett, CJ Park, Tim Strother, Nenad Tomasev, Elahe Vedadi, Ellery Wulczyn, Jan Freyberg, Roma Ruparel, Amy Wang, Renee Wong, Jonathan Krause, Christopher Semturs, David Racz, Philip Mansfield, Susan Thomas, Ewa Dominowska, Juro Gottweis, Katherine Chou, Claire Cui, Ali Eslami, Greg S. Corrado, Michael Howell, Karen DeSalvo, Jeff Dean, Zoubin Ghahramani and Demis Hassabis for their support during the course of this project.

\subsubsection*{Data Availability}
Some of the datasets used in the development of AMIE are open-source (MedQA, MultiMedQA, MIMIC-III).

\subsubsection*{Code Availability} AMIE is an LLM based research AI system for conversational disease management. We are not open-sourcing model code and weights due to the safety implications of unmonitored use of such a system in medical settings. In the interest of responsible innovation, we will be working with research partners, regulators, and providers to validate and explore safe onward uses of AMIE. For reproducibility, we have documented technical deep learning methods while keeping the paper accessible to a clinical and general scientific audience. Our work builds upon Gemini 1.5, for which technical details have been described extensively in the technical report \cite{geminiteam2024gemini15unlockingmultimodal}.

\subsubsection*{Competing Interests}
This study was funded by Alphabet Inc and/or a subsidiary thereof (‘Alphabet’). Authors who are employees of Alphabet may own stock as part of the standard compensation package. 

\newpage
\setlength\bibitemsep{3pt}
\printbibliography
\balance
\clearpage

\end{refsection}

\newpage
\begin{refsection}

\clearpage

\renewcommand{\thesection}{A.\arabic{section}}
\renewcommand{\thefigure}{A.\arabic{figure}}
\renewcommand{\thetable}{A.\arabic{table}} 
\renewcommand{\theequation}{A.\arabic{equation}} 
\renewcommand{\theHsection}{A\arabic{section}}

\setcounter{section}{0}
\setcounter{figure}{0}
\setcounter{table}{0}
\setcounter{equation}{0}

\noindent \textbf{\LARGE{Appendix}}\\
\normalfont

In the following sections, we report additional data and detailed analyses to further explain the AMIE system and results.

Mx Agent:
\begin{itemize}[leftmargin=1.5em,rightmargin=0em]
\item \cref{appendix:mx_plan} illustrates an example management plan for a patient with undiagnosed pheochromocytoma.
\item \cref{appendix:mx_auto_eval} describes the design optimization of the Mx Agent, presenting auto-evaluation of various strategies on a set of 20 validation scenarios.
\item \cref{appendix:reasoning_traces} presents examples of reasoning traces produced by the Mx Agent to produce management plans for three fictive patients.
\item \cref{appendix:mx_prompt} lists prompts used by the Mx Agent at inference time.
\end{itemize}

Dialogue Agent:
\begin{itemize}[leftmargin=1.5em,rightmargin=0em]
\item \cref{sec:datasets} details the generation and auto-evaluation of single and multi-visit simulated dialogues for fine-tuning the Dialogue Agent.
\item \cref{methods:fine_tuning} details the post-training procedure for the Dialogue Agent, including supervised fine-tuning (\cref{appendix:sft}) and reinforcement learning from human/AI feedback (\cref{appendix:rlhf}).
\item \cref{appendix:dialogue_gen_prompts} lists prompts used to generate the single-visit and simulated multi-visit doctor-patient dialogues used to fine-tune the Dialogue Agent.
\item \cref{appendix:auto_eval_filtering} lists prompts used to auto-evaluate and filter dialogues for fine-tuning the Dialogue Agent.
\item \cref{appendix:dialogue_agent_prompts} lists prompts used by the Dialogue Agent at inference time, including prompts for its chain-of-reasoning (\cref{prompts:chain_of_reasoning}) and prompts for updating the agent state (\cref{prompts:agent_state}).
\end{itemize}

OSCE Study:
\begin{itemize}[leftmargin=1.5em,rightmargin=0em]
\item \cref{appendix:rubrics} displays the MXEKF and Guidelines Use, Preciseness \& Memory evaluation rubrics used in the OSCE study. Refer to \cite{tu2024towards} for the other evaluation rubrics.
\item \cref{appendix:eval_ui} displays the rating interfaces which specialist physicians used to assess OSCE conversations and proposed management plans.
\item \cref{sec:distribution_of_patient_actor_ratings} presents the distribution of patient actor ratings for both AMIE and PCP conversations across our 100 multi-visit scenarios.
\item \cref{sec:distribution_of_specialist_physician_ratings} presents the distribution of specialist physician ratings for both AMIE and PCP conversations and management plans across our 100 multi-visit scenarios.
\end{itemize}

RxQA Benchmark:
\begin{itemize}[leftmargin=1.5em,rightmargin=0em]
\item \cref{appendix:rxqa_results} presents an example RxQA question (\cref{fig:rxqa_example_question}) and a breakdown of results for both AMIE and PCPs on this benchmark.
\item \cref{rxqa_prompts} provides more details and prompts to explain the development of the RxQA benchmark.
\end{itemize}

\clearpage
\section{Example Management Plan}
\label{appendix:mx_plan}

\begin{table}[ht!]
\footnotesize
\centering
\caption{\textbf{A management plan generated for a fictive case of undiagnosed Pheochromocytoma.} Management plans delineate the patient's clinical journey. Plans schedule both in-visit and post-visit actions. Each management item comes with generated references to guidelines for further interpretability. The plans are generated by the Mx agent based on a set of retrieved guidelines. The identical plan structure is used throughout evaluations.}
\vspace{0.5em}
\label{tab:mx_plan}
\resizebox{\textwidth}{!}{%
\begin{tabular}{p{0.15\linewidth} p{0.8\linewidth} p{0.18\linewidth}}
\toprule
\textbf{Category} & \textbf{Items} & \textbf{References} \\
\midrule
\multirow{4}{*}{\begin{tabular}[l]{@{}l@{}}\textbf{In-visit} \\ \textbf{investigations}\end{tabular} } 
& 1. Assess Gurmeet's headaches in detail: onset, duration, location, character, severity, associated symptoms (nausea, vomiting, photophobia, phonophobia), and any triggers. & -- \\
&2. Inquire about other symptoms suggestive of pheochromocytoma: postural hypotension, abdominal pain, pallor, anxiety, and changes in vision. & \texttt{bmj163} \\

& 3. Review Gurmeet's medication history, including over-the-counter pain relievers, to identify potential interactions or side effects. & -- \\

& 4. Measure Gurmeet's blood pressure in both arms, heart rate, respiratory rate, and temperature. & \texttt{ng136}, \texttt{bmj26} \\
    
\midrule
\multirow{4}{*}{\begin{tabular}[l]{@{}l@{}}\textbf{Ordered} \\ \textbf{investigations}\end{tabular} } 
& 1. Order a 12-lead ECG to assess for cardiac arrhythmias. & \texttt{bmj572} \\
& 2. Order a comprehensive metabolic panel including fasting blood glucose, HbA1c, lipid profile, and thyroid function tests (TSH, Free T4). & \texttt{bmj26}, \texttt{ng136} \\

& 3. Order plasma free metanephrines and 24-hour urine fractionated metanephrines and normetanephrines to assess for pheochromocytoma. & \texttt{bmj163}, \texttt{ng136} \\

& 4. Order a renal function test (serum creatinine, eGFR, and urine albumin:creatinine ratio). & \texttt{bmj26}, \texttt{ng136} \\
& {5. Consider abdominal CT or MRI scan if initial tests suggest pheochromocytoma.} & \texttt{bmj163} \\    
\midrule
\multirow{4}{*}{\begin{tabular}[l]{@{}l@{}}\textbf{Ordered}\\\textbf{interventions}\\\textbf{and}\\\textbf{recommendations}\end{tabular} } 
& 1. Advise Gurmeet on lifestyle modifications: reduce sodium intake, maintain a healthy diet (DASH diet), increase physical activity, and moderate alcohol consumption. & \texttt{ng136}, \texttt{bmj26} \\
& 2. Prescribe a short course of over-the-counter analgesics (e.g., paracetamol) for headache relief while awaiting test results. & -- \\

& 3. Schedule a follow-up appointment in 1-2 weeks to review test results and discuss further management. & -- \\

& 4. If pheochromocytoma is confirmed, refer Gurmeet to an endocrinologist and surgeon for specialized management. & \texttt{bmj163} \\

& 5. Educate Gurmeet about the importance of monitoring his blood pressure regularly at home  and seeking immediate medical attention if he experiences severe headaches, palpitations, or excessive sweating.  & \texttt{ng136}, \texttt{bmj26}, \texttt{bmj163} \\

\bottomrule 
\end{tabular}}
\end{table}

Management plans state the recommended sequence of actions for the patient to follow. These plans schedule future actions across short and longer term time scales, which is crucial for effective disease management. We categorize these actions into

\begin{enumerate}
\item a set of \textbf{in-visit investigations}, which are simple tests or questions that must be conducted with the patient while conversing through the online interface (e.g., measure body temperature with a home thermometer, ask about recent symptoms)

\item a set of \textbf{post-visit actions and recommendations}, which generally require physical interaction, often with healthcare providers, or lifestyle change and need to be executed or started before the next follow-up visit. The post-visit actions are themselves divided into two categories: 

\begin{enumerate}
    \item  the \textbf{ordered investigations}, which summarize a number of relevant tests requiring formal orders, such as ordering blood work or running an MRI scan (aiming to gather more diagnostic information)

    \item the \textbf{ordered interventions and recommendations}, which summarize clinical recommendations and other interventions requiring formal orders, such as prescribing a drug or even referring to surgery (representing direct therapeutic actions). 
\end{enumerate}
\end{enumerate}
For example, an in-visit investigation might be to ``ask about recent travel history'', while an ordered investigation could be ``order a chest X-ray'', and an ordered intervention might be ``prescribe an antibiotic''. To trace clinical decisions back to clinical guidelines, each plan item comes with citations to clinical guidelines. \cref{tab:mx_plan} an example of a management plan generated for a fictive case on undiagnosed Pheochromocytoma.
\clearpage
\section{Mx Agent Design and Auto-Evaluation}
\label{appendix:mx_auto_eval}

\begin{figure}[h]
    \centering
    \includegraphics[width=1.0\textwidth]{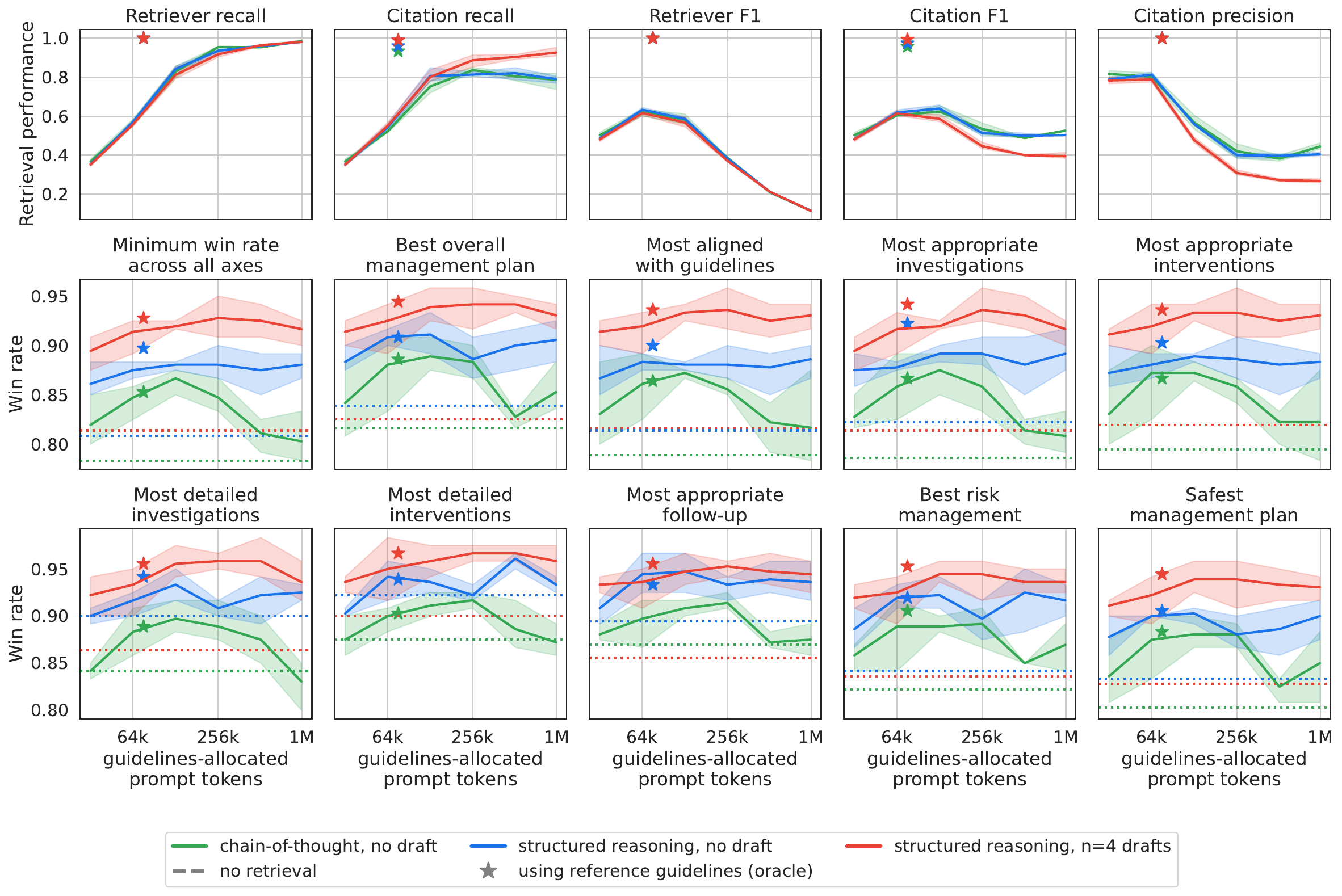}
     \vspace{0.5em}
     \caption{\textbf{Mx agent auto-evaluation.} For the selection of methods, we report validation performances aggregated over 20 evaluation scenarios and three random seeds. \textbf{Row 1.} Retrieval performances are evaluated based on the reference clinical guidelines associated with each scenario, reported based on the embedding-based retrieval step and based on the citations generated along with each management item. \textbf{Rows 2 and 3.} Management performances are measured based on the expected win-rate across 9 categories, auto-evaluated using Gemini 1.5 Pro as a judge, comparing human-written reference plans against the generated ones. We compared classic chain-of-thought (unstructured reasoning component) against our best performing reasoning structure (analyze, then state high-level goals).}
    \label{fig:mx-autoeval}
\end{figure}

We collected a separate set of 20 validation scenarios (protocol described in \cref{sec:evaluation_scenarios}), which we used as a validation set to test various Mx agent designs. Each scenario decomposes into three subsequent visits, each visit comes with a text-based case description, a list of reference guidelines, and a reference management plan. We assessed the Mx agent in isolation using, for each visit, the scenario descriptions as input and using the list of reference guidelines and the reference management plans as the prediction targets.

\subsection{Evaluation Measures} 

We measured guideline retrieval performance based on precision and recall, evaluated based on both the set of retrieved guidelines and the citations generated along with the predicted management plans. To evaluate the quality of the plan, we asked Gemini 1.5 Pro to pick a winner among the generated plan and the reference one for each of the nine following categories: \textit{best overall management plan}, \textit{best guidelines alignment}, \textit{most appropriate investigations}, \textit{most appropriate interventions}, \textit{most detailed investigations}, \textit{most detailed interventions}, \textit{most appropriate follow-up}, \textit{best risk management} and \textit{safest management plan}. The win-rate was estimated for each category by randomly shuffling the predicted and target plans in the prompt (n=4). The full auto-evaluation prompt is presented at the end of this section. We report metrics with in-context reference guidelines and with a range of in-context retrieved guidelines, measured by their total number of tokens. We acknowledge the limitations of this evaluation tool due to self-enhancement and verbosity  biases~\cite{zheng2023judgingllmasajudgemtbenchchatbot, panickssery2025llm}. The Mx auto-evaluator prompt is provided in \cref{prompt:mx_autoeval}.

\subsection{Agent Designs} We explored diverse reasoning structures, ensemble refinement \cite{singhal2023towards} and varying numbers of context lengths. We explored reasoning structures varying from simple chain-of-thought \cite{wei2022chain, kojima2022large} to more complex reasoning trees designed to force the model to interleave in-context retrieval with reasoning. Our search space is limited by serving constrains, which are the maximum number of parallel model calls (set to four) and a response time (targeted to approximately one minute).

\paragraph{Implementation details}

When using ensemble refinement, we generate plans stochastically without conditioning on retrieved documents. This allows for concurrently retrieve and draft, reducing the overall response time. The drafts are refined into a final plan using a final model call, conditioned on the full text of the retrieved guidelines.

When retrieving guidelines based on text embedding, for each query, we separately rank clinical guideline sources based on text embeddings. The final list of guidelines is obtained by iteratively taking the next highest-ranked guideline from each available guideline corpus until the maximum token limit is reached.

\subsection{Results}

\cref{fig:mx-autoeval} summarizes the retrieval and auto-evaluated management metrics for a subset of methods: a simple chain-of-thought method without drafting, structured reasoning without drafting, and structured reasoning with drafting (see \cref{fig:reasoning_structures} for the definition of the reasoning structure).

\paragraph{Retrieval Performance}  The coarse retrieval step (query generation + text-embeddings) is sufficient to score a recall superior to 90\% using a context size of 256k. Recall slightly decreases when measured based on generated citations. However, the F1 score measured based on citations is higher for larger context sizes than it is when measured on the coarse retrieval step, suggesting that our language model discriminates part of the irrelevant guidelines via in-context processing. Furthermore, it is important to note that retrieval performances are measured based on reference guidelines, which only represent one valid set of guidelines. Therefore precision and F1 score must be interpreted with care as the rate of false positives is likely over-estimated.

\paragraph{Management Reasoning Performance}  Across all evaluation axes, Gemini favors the generated management plans over the reference ones with an average win-rate ranging from 75\% to 95\% for all methods. This corroborates previous observation that auto-evaluators favor synthetic data~\cite{zheng2023judgingllmasajudgemtbenchchatbot, panickssery2025llm}. In spite of this bias, the win-rate allows us to discern several trends. Firstly, guidelines retrieval clearly suggests an improvement in management performance. Secondly, structured reasoning improves over a classic chain-of-thought strategy. Thirdly, drafting plans (ensemble refinement) yields an additional performance boost. Fourthly, management quality scales with context size up to a critical context size that differs depending on the choice of method. Last, we measured improvements over the oracle (feeding the reference guidelines), both suggesting that non-reference guidelines are helpful and that retrieval performance improves past 64k tokens.

\subsection{Selected Design} Based on this analysis, and after qualitatively reviewing five samples with clinicians, we identified the optimal configuration: using drafts and structuring the model output as detailed in \cref{prompt:mx_plan_constraint}, which scales best with context size and leads to best results across all axes. 

Based on the minimum win rate across all axes, we fixed the context size to 256k tokens. We found that, based on qualitative assessments, even if the drafts are not generated based on the retrieved documents, ensemble methods allow generating more comprehensive plans, covering a wider range of often complementary care and investigation strategies. 

Running a single structured plan generation step takes around 50 seconds for a context size of 256k tokens, and around 30 seconds without in-context documents. A full Mx agent call, that chains retrieval, four draft generations and a final refinement step takes around 80 seconds.

\begin{figure}[h]
\resizebox*{!}{0.9\textheight}{%
\begin{minipage}{1.5\textwidth}
\begin{promptbox2}{Mx plan auto-evaluation.}
\footnotesize
\input{prompts/mx_autoeval}
\end{promptbox2}
\end{minipage}
}
\vspace{0.1cm}
\caption{\textbf{Mx plan auto-evaluation.} Assesses relative management reasoning performances using LLM as a judge. \texttt{case\_presentation} is a full-text description of the fictive scenario, written by experts. \texttt{plan\_a} and \texttt{plan\_b} are either a generated plan, or the expert-written reference plan, in random order. The \texttt{guidelines} variable corresponds to the full-text of the reference guidelines accompanying that scenario. The variable \texttt{analysis} is an expert-written paragraph of text supporting the choice of management plan and alignment with guidelines. }
\label{prompt:mx_autoeval}
\end{figure}
\clearpage
\section{Reasoning Traces and Managements Plans based on Evaluation Scenarios}\label{appendix:reasoning_traces}

\begin{figure}[h]
    \centering
    \caption{\textbf{Mx Agent reasoning trace and plan. A fictive patient treated for pheochromocytoma.} In this example, Mx Agent considers appropriate post-surgical follow up, including counseling about unlikely long-term side effects, while also following recommendations for blood pressure control, healthy diet, and smoking cessation.  }
    \label{fig:mx_sample_1}
    \vspace{0.2em}
    \includegraphics[width=0.7\linewidth]{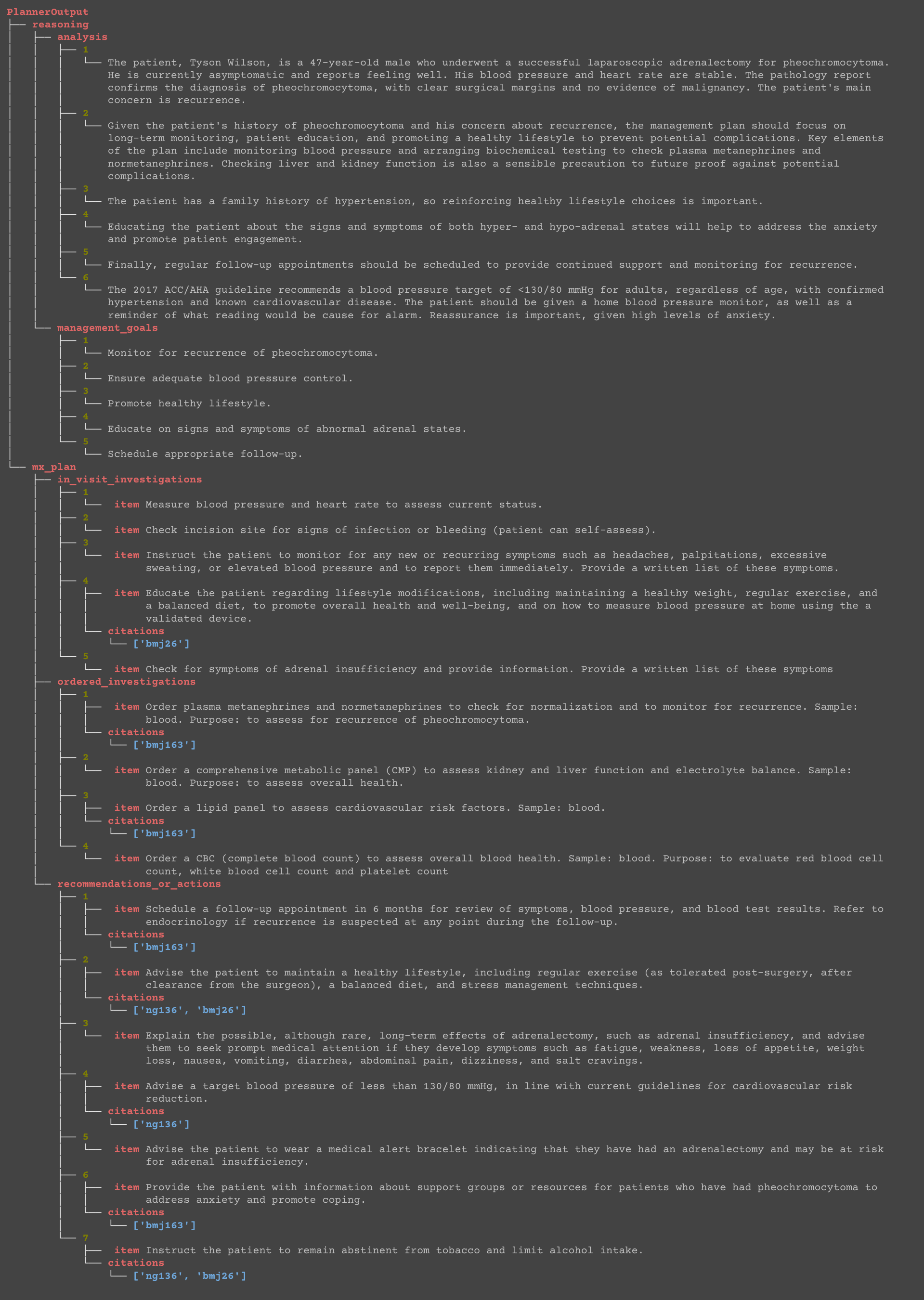}
\end{figure}

\begin{figure}[h]
    \centering
    \vspace{-1cm}
    \caption{\textbf{Mx Agent reasoning trace and plan. A fictive patient with COPD.} In this example, Mx Agent reviews the medications for each of the patient's chronic medical conditions, considering possible drug interactions (such as beta blockers worsening COPD symptoms). It also recommends appropriate cancer screening and vaccinations given his smoking history and comorbidities.}
    \label{fig:mx_sample_2}
    \vspace{0.2em}
    \includegraphics[width=0.7\linewidth]{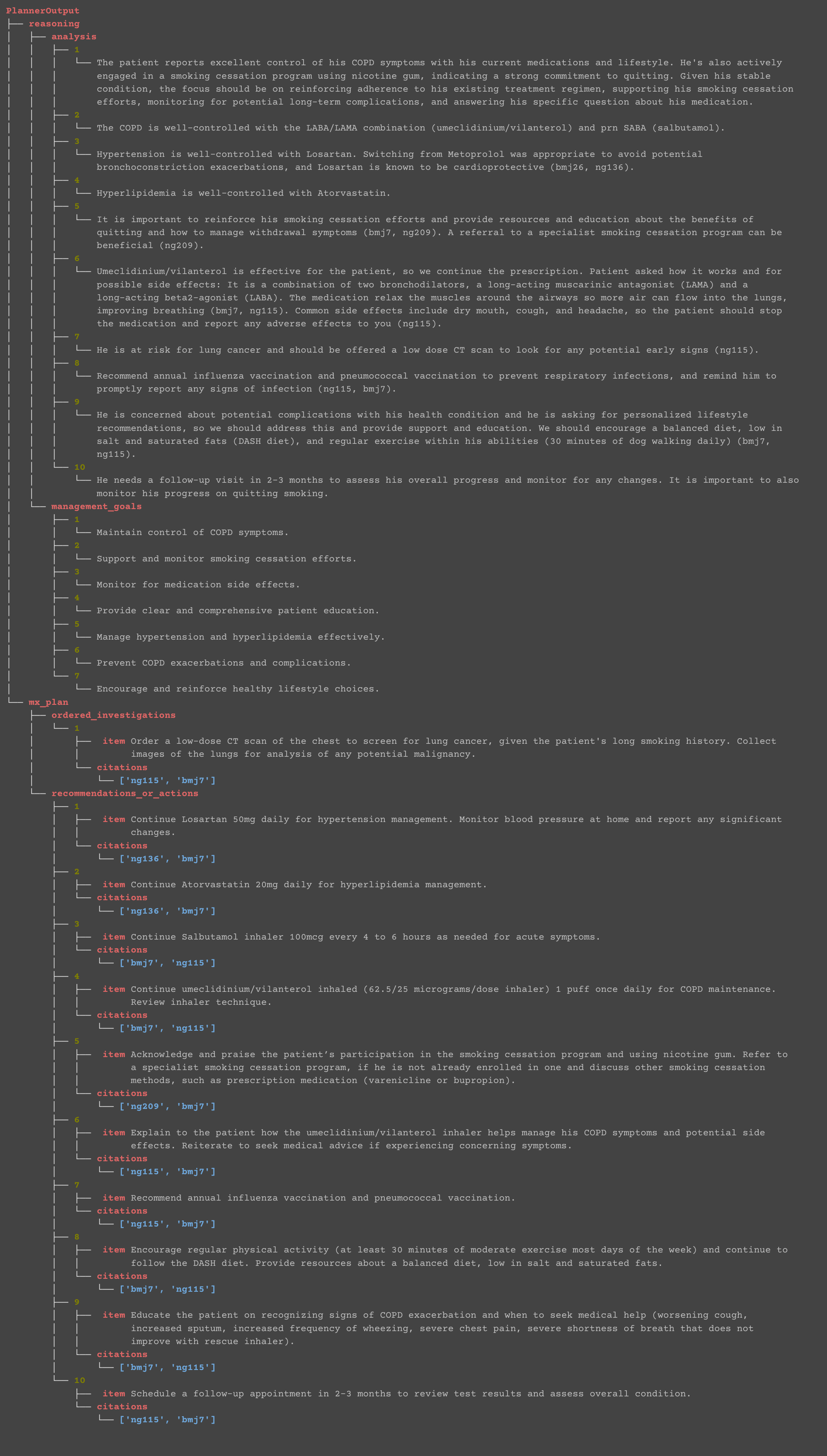}

\end{figure}

\begin{figure}[h]
    \centering
     \vspace{-1cm}
    \caption{\textbf{Mx Agent reasoning trace and plan. A fictive patient being followed up for a major depressive disorder.} In this example, the Mx Agent develops an expectant plan for major depressive disorder that combines medication, lifestyle changes, and cognitive behavioral therapy, with appropriate monitoring and follow up, including a possible tapering plan.}
    \label{fig:mx_sample_3}
   \vspace{0.2em}
    \includegraphics[width=0.7\linewidth]{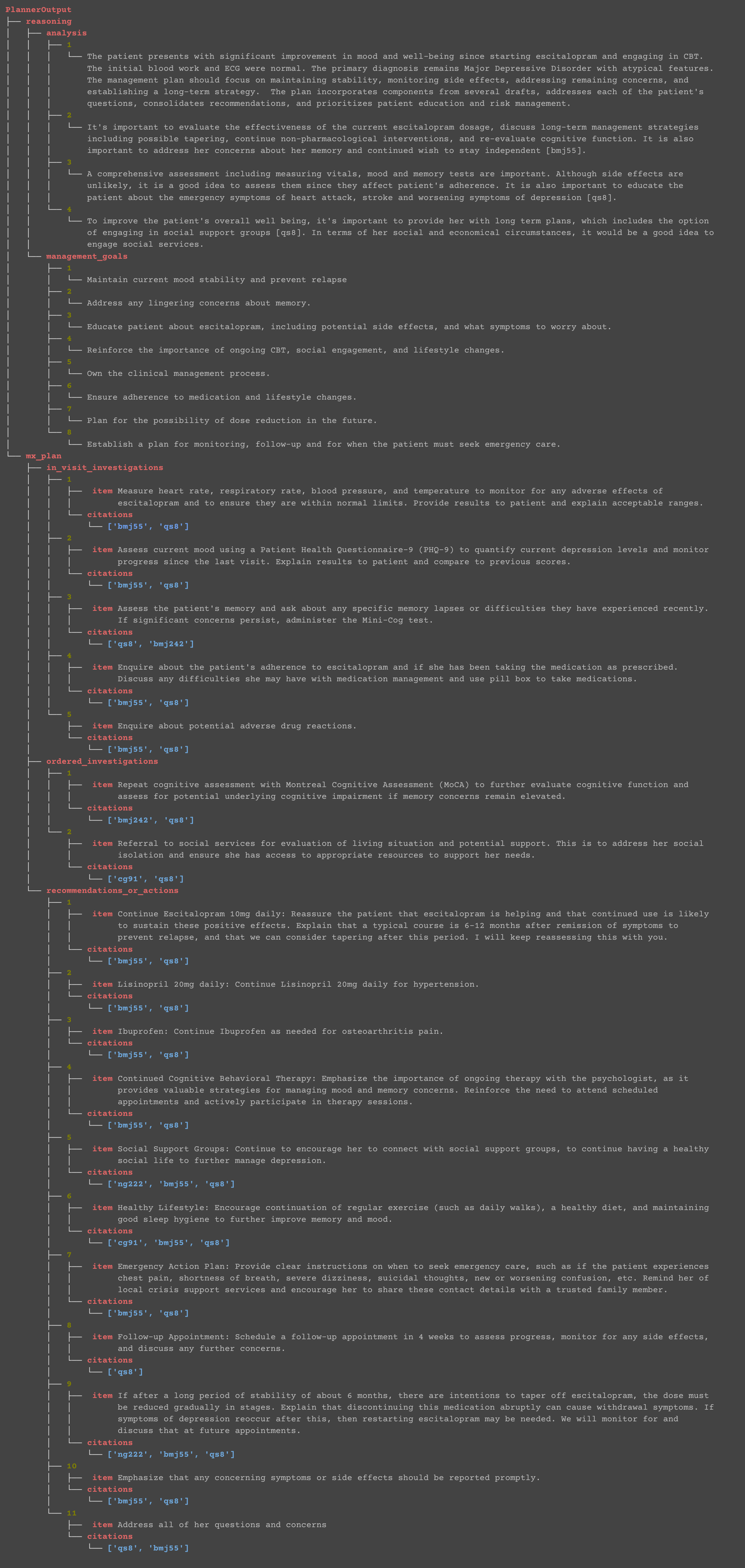}
   
\end{figure}

This section presents a selection of model outputs, converted from JSON to trees for readability and displayed using the \texttt{rich} Python library.\footnote{\url{https://github.com/Textualize/rich}} We present model outputs based on fictive case scenarios used in \cref{appendix:mx_auto_eval}:
\begin{enumerate}
    \item \cref{fig:mx_sample_1}: a fictive patient treated for Pheochromocytoma.
    \item \cref{fig:mx_sample_2}: a fictive patient a history of COPD.
    \item \cref{fig:mx_sample_3}: a fictive patient being followed up for a major depressive disorder.
\end{enumerate}

\clearpage
\section{Mx Agent Prompts}
\label{appendix:mx_prompt}

In this section, we present the prompts and decoding constraints used by the Mx agent (\cref{sec:mx_agent}):
\begin{itemize}
    \item \cref{prompt:mx_abstract_gen} is the prompt and constraint utilized to preprocess reference documents prior to retrieval.
    \item \cref{prompt:mx_query_gen} us the prompt and constraint utilized to generate retrieval queries.
    \item \cref{prompt:mx_plan} is the main prompt, used to reason and generate structured management plans.
    \item \cref{prompt:mx_plan_merge} a suffix to the plan generation prompt, appended when refining drafts.
    \item \cref{prompt:mx_plan_constraint} the decoding constraint applied for plan generation.
\end{itemize}

\begin{figure}[h]
\begin{promptbox2}{Mx agent prompt: generate document titles and abstracts.}
\footnotesize
\input{prompts/mx_abstract_gen.tex}
\end{promptbox2}
\vspace{0.1cm}
\caption{\textbf{Mx Agent prompt:} generate document titles and abstracts.}
\label{prompt:mx_abstract_gen}
\end{figure}

\begin{figure}[h]
\begin{promptbox2}{Mx agent prompt: generate search queries.}
\footnotesize
\input{prompts/mx_query_gen.tex}
\end{promptbox2}
\vspace{0.1cm}
\caption{\textbf{Mx Agent prompt:} generate search queries.}
\label{prompt:mx_query_gen}
\end{figure}

\begin{figure}[h]
\resizebox*{!}{\textheight}{%
\begin{minipage}{1.5\textwidth}
\begin{promptbox2}{Mx agent prompt: generating management plans.}
\footnotesize
\input{prompts/mx_plan.txt}
\end{promptbox2}
\end{minipage}
}
\vspace{0.1cm}
\caption{\textbf{Mx Agent prompt:} generate management plan.}
\label{prompt:mx_plan}
\end{figure}

\begin{figure}[h]
\begin{promptbox2}{Mx agent prompt: refine management plans.}
\footnotesize
\input{prompts/mx_plan_merge}
\end{promptbox2}
\vspace{0.1cm}
\caption{\textbf{Mx Agent prompt:} refine management plan.}
\label{prompt:mx_plan_merge}
\end{figure}

\begin{figure}[h]
\begin{promptbox2}{Mx agent: structural constraint.}
\footnotesize
\input{prompts/mx_plan_constraint}
\end{promptbox2}
\vspace{0.1cm}
\caption{\textbf{Mx Agent:} structural constraint. Decoding constraints allow generating structurally consistent plans and enforcing citations that always match the set of retrieved guidelines.}
\label{prompt:mx_plan_constraint}
\end{figure}

\clearpage
\section{Development of Training Dialogues}
\label{sec:datasets}
                                 
\subsection{Augmenting Single-Visit Dialogues}\label{par:real_world_dialogue}
The original PaLM-2-based version of AMIE described in~\citet{tu2024towards} was trained with both real-world dialogue transcripts and simulated doctor-patient dialogues. The real-world transcripts were particularly noisy, containing frequent interruptions, paraverbal annotations (like `[laughing]'), and many details which were clinically irrelevant or unsuitable for a text-based setting (like physical examination). While generally higher quality, the simulated dialogues also had issues such as a lack of recommendations to the patient. To resolve these issues, we used Gemini 1.5 Pro to rewrite the original dialogues. Furthermore, while the inclusion of follow-up visits is useful for our multi-visit setting, the follow-up visits in the original dataset lacked any context about what had transpired in prior visits. To solve this, we used Gemini 1.5 pro to generate context describing the doctor's prior knowledge about the patient, which was unique per dialogue and provided in the instructions during supervised fine-tuning (see \cref{methods:fine_tuning}). Prompting details for this process are listed in~\cref{sec:real_world_dialogue_prompts}.

\subsection{Simulated Longitudinal Interactions}\label{methods:synth_data}

\begin{figure}[ht!]
    \centering
    \includegraphics[width=0.6\textwidth]{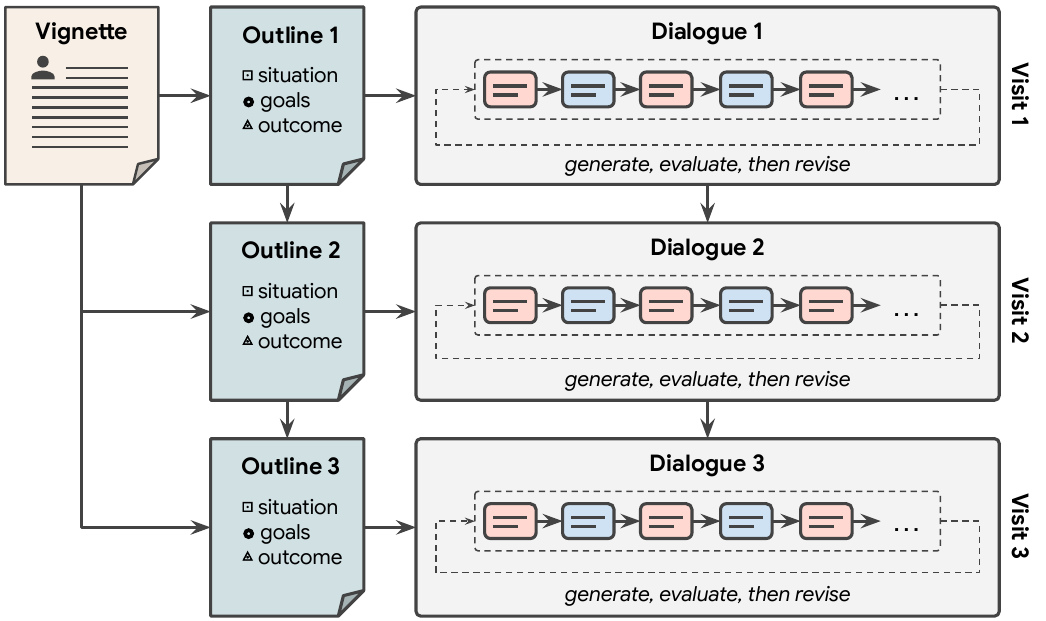}
    \vspace{0.4cm}
    \caption{\textbf{Top-down longitudinal dialogue synthesis.} We generate multi-visit doctor-patient interactions based using a top-down approach: first generating a high-level patient vignette, then generating an outline for each visit, finally by generating the doctor-patient dialogue for each visit. Each dialogue is drafted, then auto-evaluated and revised.}
    \label{fig:simulated_mx_dialogue}
\end{figure}

We expanded the training set of patient-doctor dialogues with synthetic dialogues simulating the clinical course of patients from their initial presentations to their resolutions. Each simulated dialogue comes with a fictive patient vignette, a patient-doctor dialogue for a maximum of three subsequent visits, and reports summarizing the outcome of the interventions (e.g., surgery) and investigations (e.g., blood work) ordered during the preceding visit, if any. 

Inspired by the prior work on co-writing screenplays from \citeauthor{mirowski2023co}~\cite{mirowski2023co}, we designed a top-down synthesis pipeline, starting from a high-level patient vignette to goal-directed visit outlines, and finally down to the dialogue level for each outline (illustrated in \cref{fig:simulated_mx_dialogue}). Building on our prior work \cite{tu2024towards}, we generated a wide range of synthetic patients vignettes grounded in publicly available knowledge retrieved using Google Search. For each vignette, we generated visit outlines one after another for all three visits. Each outline describes the situation at the start of the visit (patient state, including external reports such as lab results or radiology report), defines conversation goals for both the patient and doctor (i.e., patient expectations, doctor strategy) and states the outcome of the visit (i.e., learned patient facts, updated differential diagnosis, follow-up schedule, and ordered post-visit investigations and interventions). Once the visit outlines were generated, for each visit, we drafted the entire patient-doctor conversation using a single model call, which we then automatically evaluated and revised, akin to \citet{madaan2023self}, based on a selection of high-level criteria (communication, coherence, reasoning, and management quality). All scenarios and dialogues were generated using Gemini 1.5 Pro and constrained decoding \cite{koo2024automata}, which enabled generating complex data structures (vignettes, outlines, dialogues, evaluations) using fewer model calls, significantly increasing the pipeline efficiency. Prompting details for this process are listed in \cref{methods:synth_data}.

\subsection{Filtering Dialogues via Auto-Evaluation.}
To ensure quality of the synthetic dialogues, we leveraged LLM-based auto-evaluation to assess the quality of the doctor messages in the dialogues across various criteria with regard to hallucinations and bias, ethical behavior, quality of history taking, providing information, building rapport and empathy, and so on. We only included the dialogues which passed this set of auto-evaluation criteria in the subsequent instruction fine-tuning stage. 
Specifically, the auto-evaluation was performed using a four-criteria system:
\begin{itemize}
    \item \textbf{Undesirable doctor agent behavior:} This includes providing unsafe, biased, or non-evidence-based statements; exhibiting ethical or safety concerns; inadequately differentiating normal versus unhealthy cases; prematurely offering diagnoses without sufficient history-taking (except in emergencies); and incorrectly suggesting the ability to prescribe medications.
    \item \textbf{Desirable doctor agent behavior:} Similar to the above mentioned criteria but we inversely evaluated the criteria, ensuring a comprehensive capture of potential issues by also focusing on what constitutes positive and appropriate doctor behavior.
    \item \textbf{Hallucination and confabulation:} This criterion focuses on whether the doctor agent introduces information not explicitly or implicitly provided by the patient, mistakenly assumes or recalls information, misremembers/misunderstands/misinterprets patient input, claims access to external databases or services, or mentions non-existent medications.
    \item \textbf{Dialogue formatting:} This checks for question or information repetition, the balance of turns between patient and doctor, and adherence to minimum and maximum turn limits of the dialogue.
\end{itemize}
Detailed auto-evaluation prompts are listed in the Appendix~\cref{appendix:auto_eval_filtering}. For each of these criteria, a binary result is generated. Any dialogue receiving an unfavorable result in any category was filtered out, while the rest were retained and utilized for fine-tuning. The auto-evaluation process was done by Gemini 1.5 Pro with the constrained decoding framework~\citep{liu2024we}.

\section{Dialogue Agent Post-training}\label{methods:fine_tuning}

\subsection{Supervised Fine-tuning}\label{appendix:sft} We fine-tuned on top of the base model Gemini 1.5 Flash to adapt it for conversational and longitudinal patient interactions. The medical question-answering and summarization tasks were set up as in \citeauthor{tu2024towards}~\cite{tu2024towards}, where task-specific instructions were prepended to each question-completion pair.

For single-visit dialogues, AMIE was trained to predict the next conversational turn based on all previous turns, assuming the role of the doctor. Here, we prepended instructions that described the task of interviewing the patient. Furthermore, for follow-up visits, our instructions contained scenario-specific information which was unique for each dialogue and was generated to contextualize the visit (see \cref{par:real_world_dialogue}).

Finally, for the longitudinal multi-visit dialogues described in 
\cref{methods:synth_data}, we concatenated the sequence of visits into a single multi-visit transcript. Between each visit, we inserted the inter-visit context, such as reports from test results or updates which would be available to the doctor during the following visit. We again sampled doctor turns as targets, training AMIE to predict the next conversational turn based on the preceding multi-visit dialogue history.

\subsection{Reinforcement Learning}\label{appendix:rlhf} To further align the model's outputs with the desired preferences on conversational diagnostic and management reasoning, we employed reinforcement learning from feedback~\citep{ouyang2022training}, with a hybrid of both human and LLM preferences. The process includes two phases---reward model training, and reinforcement learning. The reward model predicts how a rater (human or LLM) would rate a given response, which is then used in the reinforcement learning stage such that the policy model (the dialogue agent) will be optimized to generate outputs that receive high scores judged by the reward model, thus indirectly aligning with the rater preferences.

We first trained a Gemini 1.5 Flash-based reward model using clinically relevant pairwise preference data from two distinct sources:
\begin{itemize}
    \item \textbf{Preferences from human experts:} We curated three pairwise preference datasets annotated by clinical experts including: (1) Single-visit dialogue response preference task, (2) AMIE cardiology case evaluation task~\citep{o2024towards}, and (3) AMIE breast cancer case evaluation task~\citep{palepu2024exploring}.
    \item \textbf{Preferences from LLM:} Three pairwise preference datasets rated by LLM were curated where we utilized Gemini 1.5 Pro as an auto-rater to provide AI-generated preferences between pairs of generated responses, including: (1) Single-visit dialogue response preference task (Gemini 1.5 Pro acting as the judge), (2) MedQA medical reasoning task, and (3) electronic health record (EHR) summarization task.
\end{itemize}

Following the supervised fine-tuning, we next refined the dialogue agent (policy model) via reinforcement learning using the reward signal provided by the previously trained reward model. For this phase, we utilized (1) a separate set of single-visit dialogue examples similar to the ones used in the SFT stage, along with (2) the MedQA medical reasoning examples~\citep{jin2021disease}. The process proceeded as follows: a prompt was sampled from this dataset and presented to the policy model to generate a response. The reward model evaluated this response, assigning a scalar reward value. Finally, the reward signal was used to update the parameters of the policy model via the reinforcement learning algorithm used in Gemini 1.5~\citep{geminiteam2024gemini15unlockingmultimodal}. This process allowed the dialogue agent to learn to generate responses that maximized the reward, leading to improved performance. Further details of the data used for reward model training and reinforcement learning are listed in~\cref{appendix:rm_data} and \cref{appendix:rl_data}.
\subsection{Dataset information for reinforcement learning from feedback}

\begin{table}[ht!]
\centering
\resizebox{\textwidth}{!}{
\begin{tabular}{cclccc}
\toprule
\textbf{Task} & \textbf{Sample size} & \multicolumn{1}{c}{\textbf{Description}} & \textbf{Response 1} & \textbf{Response 2} & \textbf{Rater} \\
\midrule
\begin{tabular}[c]{@{}c@{}}Single visit dialogue \\ response preference (human)\end{tabular} & 470 & \begin{tabular}[c]{@{}l@{}}A case vignette, a generated dialogue up to the last user turn,\\ and a pair of responses are given to the rater to decide the \\ preference. All dialogues are derived from 464 common \\ medical conditions.\end{tabular} & AMIE & Gemini 1.5 Pro & Human \\
\midrule
\begin{tabular}[c]{@{}c@{}}Cardiology case \\ management preference\end{tabular} & 1473 & \begin{tabular}[c]{@{}l@{}}A cardiology case, and a pair of management recommendations.\\ Note that the AMIE responses had been rewritten to the same \\ style as the expert responses since the expert responses are \\ shorter and more specific. The expert response is by default the \\ better response.\end{tabular} & AMIE & Human & Human \\
\midrule
\begin{tabular}[c]{@{}c@{}}Breast cancer case \\ management preference\end{tabular} & 100 & \begin{tabular}[c]{@{}l@{}}An oncology case, and a pair of management recommendations. \\ The expert response is by default the better response.\end{tabular} & AMIE & Human & Human \\
\midrule
\begin{tabular}[c]{@{}c@{}}Single visit dialogue \\ response preference (LLM)\end{tabular} & 2184 & \begin{tabular}[c]{@{}l@{}}A case vignette, a generated dialogue until the last user turn, \\ and a pair of responses. An LLM rater outputs a preference \\ score for each response, ranging from 1 to 7. Dialogues are \\ derived from 464 common medical conditions and 628 \\ uncommon conditions.\end{tabular} & \begin{tabular}[c]{@{}c@{}}AMIE\\ Gemini 1.5 Pro\end{tabular} & Gemini 1.5 Flash & Gemini 1.5 Pro \\
\midrule
\begin{tabular}[c]{@{}c@{}}MedQA medical \\ reasoning preference\end{tabular} & 191 & \begin{tabular}[c]{@{}l@{}}A MedQA question and the corresponding answer, and a pair \\ of answer explanations. An LLM rater outputs a preference \\ score for each response, ranging from 1 to 7.\end{tabular} & MedPaLM 2 & MedPaLM 2 & Gemini 1.5 Pro \\
\midrule
\begin{tabular}[c]{@{}c@{}}EHR summarization\\ preference\end{tabular} & 65 & \begin{tabular}[c]{@{}l@{}}A clinical text, and a pair of summarized texts. An LLM rater \\ outputs a preference score for each response, ranging from 1 to 7.\end{tabular} & MedPaLM 2 & MedPaLM 2 & Gemini 1.5 Pro \\
\bottomrule
\end{tabular}
}
\vspace{0.1cm}
\caption{\textbf{Pairwise preference datasets for reward model training.} Preference data mixture for training the reward model.}
\label{appendix:rm_data}
\end{table}

\begin{table}[ht!]
\centering
\resizebox{\textwidth}{!}{
\begin{tabular}{ccl}
\toprule
\textbf{Task} & \textbf{Sample size} & \multicolumn{1}{c}{\textbf{Description}} \\
\midrule
\begin{tabular}[c]{@{}c@{}}Single-visit dialogue \\ response generation\end{tabular} & 13104 & \begin{tabular}[c]{@{}l@{}}Given a case vignette and a generated dialogue until the last user turn. The prompt asks the model to play the \\ medical expert role and provide the appropriate response for the prior turns. All dialogues are derived from \\ 464 common medical conditions and 628 uncommon conditions.\end{tabular} \\
\midrule
\begin{tabular}[c]{@{}c@{}}MedQA medical reasoning \\ explanation preference\end{tabular} & 191 & \begin{tabular}[c]{@{}l@{}}Given a MedQA question and the corresponding answer, the prompt asks the model to generate an explanation \\ of the medical reasoning of the answer.\end{tabular} \\
\bottomrule
\end{tabular}
}
\vspace{0.1cm}
\caption{\textbf{Datasets for reinforcement learning.}}
\label{appendix:rl_data}
\end{table}
\clearpage
\section{Prompts for Dialogue Generation}
\label{appendix:dialogue_gen_prompts}
\subsection{Single-Visit Dialogue Rewrite Prompts}
\label{sec:real_world_dialogue_prompts}

The single-visit dialogues initially introduced in \citet{tu2024towards} were revised by Gemini 1.5 Pro to improve their quality and contextualize them (see \cref{sec:datasets}).
\begin{enumerate}
    \item For real world dialogues, an outline of the original transcript is generated, with modifications made to adapt it to a virtual, text-based setting (\cref{prompt:real_dialogue_outline}).
    \item Gemini revises each dialogue to improve its suitability for instruction fine-tuning. For real-world transcripts, Gemini leverages the outline generated in the prior step to structure its revision (\cref{prompt:real_dialogue_rewrite}).
    \item For each revised dialogue, Gemini generates fictive context detailing the information which was known to the doctor prior to that conversation, which is utilized during supervised fine-tuning (\cref{prompt:real_dialogue_scenario}).
\end{enumerate}

\begin{figure}[h!]
\begin{promptbox2}{Reorganizing Real-world Dialogue Transcripts}
\footnotesize
\input{prompts/sv_dialogue_outline.txt}
\end{promptbox2}
\vspace{0.1cm}
\caption{\textbf{Single-Visit Dialogue Rewrite.} Reorganizing Real-world Dialogue Transcripts}
\label{prompt:real_dialogue_outline}
\end{figure}

\begin{figure}[h!]
\begin{promptbox2}{Revision of Single-Visit Dialogues}
\footnotesize
\input{prompts/sv_dialogue_rewrite.txt}
\end{promptbox2}
\vspace{0.1cm}
\caption{\textbf{Single-Visit Dialogue Rewrite.} Revision of Dialogues}
\label{prompt:real_dialogue_rewrite}
\end{figure}

\begin{figure}[h!]
\begin{promptbox2}{Generation of Doctor Context}
\footnotesize
\input{prompts/sv_dialogue_scenario.txt}
\end{promptbox2}
\vspace{0.1cm}
\caption{\textbf{Single-Visit Dialogue Rewrite.} Generation of Doctor Context}
\label{prompt:real_dialogue_scenario}
\end{figure}

\subsection{Simulated Multi-visit Dialogue Generation Prompts} \label{prompts:mx_dialogues}

We simulate multi-visit dialogues by first generating $N$ visit outlines using the prompt presented in \cref{prompt:multi_visit_dialogue_outline} based on patient vignettes produced in \cite{tu2024towards}.
Given the generated outlines, we generate the dialogues visit by visit. For each visit, we generated a dialogue draft using a single model call using the prompt presented in \cref{prompt:multi_visit_dialogue_generation}, which we then auto-evaluate using \cref{prompt:multi_visit_dialogue_critique}, and refine using \cref{prompt:multi_visit_dialogue_generation} again.

\begin{figure}[h]
\resizebox*{!}{0.8\textheight}{%
\begin{minipage}{1.4\textwidth}
\begin{promptbox2}{Generating dialogue outlines}
\footnotesize
\input{prompts/multi_visit_dialogue_outline.tex}
\end{promptbox2}
\end{minipage}
}
\vspace{0.1cm}
\caption{\textbf{Multi-visit Dialogue Generation.} Generating multi-visit dialogue outlines.}
\label{prompt:multi_visit_dialogue_outline}
\end{figure}

\begin{figure}[h]
\resizebox*{!}{\textheight}{%
\begin{minipage}{1.4\textwidth}
\begin{promptbox2}{Generating or improving dialogues}
\footnotesize
\input{prompts/multi_visit_dialogue_generation.tex}
\end{promptbox2}
\end{minipage}
}
\vspace{0.1cm}
\caption{\textbf{Multi-visit Dialogue Generation.} Generating or improving dialogues based on critique.}
\label{prompt:multi_visit_dialogue_generation}
\end{figure}

\begin{figure}[h]
\begin{promptbox2}{Multi-visit Dialogue Auto-Evaluation}
\footnotesize
\input{prompts/multi_visit_dialogue_critique}
\end{promptbox2}
\vspace{0.1cm}
\caption{\textbf{Multi-visit Dialogue Generation.} Auto-evaluating multi-visit dialogues.}
\label{prompt:multi_visit_dialogue_critique}
\end{figure}
\clearpage
\section{Auto-Evaluation for Dialogue Filtering}\label{appendix:auto_eval_filtering}

\begin{figure}[h]
\begin{promptbox2}{Auto-eval for dialogue filtering (hallucination)}
\footnotesize
\input{prompts/autoeval_filter_hallucination.tex}
\end{promptbox2}
\vspace{0.1cm}
\caption{\textbf{Prompt: auto-eval for dialogue filtering (hallucination)}}
\label{fig:autoeval_filter_hallucination}
\end{figure}

\begin{figure}[h]
\begin{promptbox2}{Auto-eval for dialogue filtering (doctor agent behavior)}
\footnotesize
\input{prompts/autoeval_filter_behave.tex}
\end{promptbox2}
\vspace{0.1cm}
\caption{\textbf{Prompt: auto-eval for dialogue filtering (behaviour)}}
\label{fig:autoeval_filter_behave}
\end{figure}

\clearpage
\section{Dialogue Agent Prompts}
\label{appendix:dialogue_agent_prompts}

\subsection{Chain-of-reasoning Prompts}\label{prompts:chain_of_reasoning}

The Dialogue Agent uses 3 sequential prompts to generate responses to patient messages. These prompts each leverage the current dialogue history, the current agent state, including the patient summary, differential diagnosis, and management plan, and change during follow-up visits with a patient.

The first step in the chain of reasoning is to analyze the context and generate a current plan for what is needed next in the conversation (\cref{fig:chain_of_reasoning_prompt_1}). For example, the output of this step might be that the agent should ask about family history if that information is relevant/missing, or alternatively, it may be that the agent should wrap up the conversation if it feels ready to do so. The second step in the chain-of-reasoning is to execute its plan, and generate a response to the patient (\cref{fig:chain_of_reasoning_prompt_2}). The final step in the chain-of-reasoning is to refine its drafted response prior to sending it out (\cref{fig:chain_of_reasoning_prompt_3}). This is useful because there are several potential failure modes that we would like to avoid with responses, such as asking questions that have already been asked, and an additional prompt helps mitigate these.

\begin{figure}[h!]
\begin{promptbox2}{Dialogue Agent - Plan Response}
\footnotesize
\input{prompts/dialogue_agent_plan_response}
\end{promptbox2}
\vspace{0.1cm}
\caption{\textbf{Planning - First prompt in the dialogue agent chain-of-reasoning.}}
\label{fig:chain_of_reasoning_prompt_1}
\end{figure}

\begin{figure}[h!]
\begin{promptbox2}{Dialogue Agent - Generate Response}
\footnotesize
\input{prompts/dialogue_agent_generate_response}
\end{promptbox2}
\vspace{0.1cm}
\caption{\textbf{Generation - Second prompt in the dialogue agent chain-of-reasoning.}}
\label{fig:chain_of_reasoning_prompt_2}
\end{figure}

\begin{figure}[htbp!]
\begin{promptbox2}{Dialogue Agent - Refine Response}
\footnotesize
\input{prompts/dialogue_agent_refine_response}
\end{promptbox2}
\vspace{0.1cm}
\caption{\textbf{Refinement - Third prompt in the dialogue agent chain-of-reasoning.}}
\label{fig:chain_of_reasoning_prompt_3}
\end{figure}
\clearpage
\subsection{Agent State Prompts}\label{prompts:agent_state}

The dialogue agent state is periodically updated over the course of a conversation. While the management plan updates are performed by the separate Mx Agent (see \cref{sec:mx_agent}), the running patient summary and differential diagnosis updates are handled by the Dialogue Agent itself. The prompt for the patient summary update is shown in \cref{fig:patient_summary_update_prompt}, while the differential diagnosis update (DDx) prompt is shown in \cref{fig:ddx_update_prompt}.

\begin{figure}[h!]
\begin{promptbox2}{Dialogue Agent - Patient Summary Update}
\footnotesize
\input{prompts/dialogue_agent_state_prompt}
\end{promptbox2}
\vspace{0.1cm}
\caption{\textbf{Patient Summary Update Prompt.}}
\label{fig:patient_summary_update_prompt}
\end{figure}

\begin{figure}[h!]
\begin{promptbox2}{Dialogue Agent - DDx Update}
\footnotesize
\input{prompts/dialogue_ddx_update.tex}
\end{promptbox2}
\vspace{0.1cm}
\caption{\textbf{DDx Update Prompt.}}
\label{fig:ddx_update_prompt}
\end{figure}

\clearpage

\clearpage
\section{OSCE Evaluation Rubrics}
\label{appendix:rubrics}

\begin{table}[hbt!]
    \centering
    \includegraphics[width=\textwidth,height=\textheight,keepaspectratio]{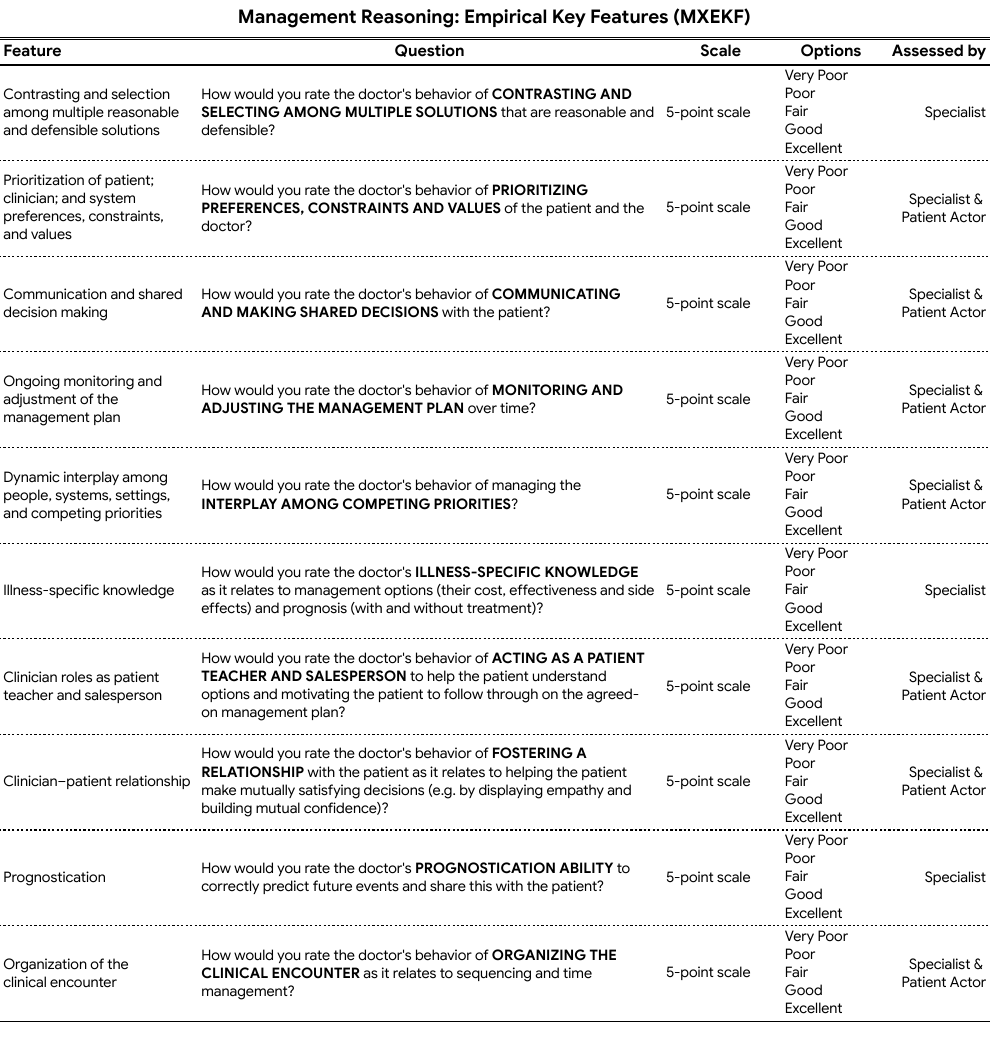}
    \caption{\textbf{Management Reasoning Empirical Key Features (MXEKF) rubric details.}}
    \label{tab:mxekf_rubric_details}
\end{table}

\begin{table}[hbt!]
    \centering
    \includegraphics[width=\textwidth,height=\textheight,keepaspectratio]{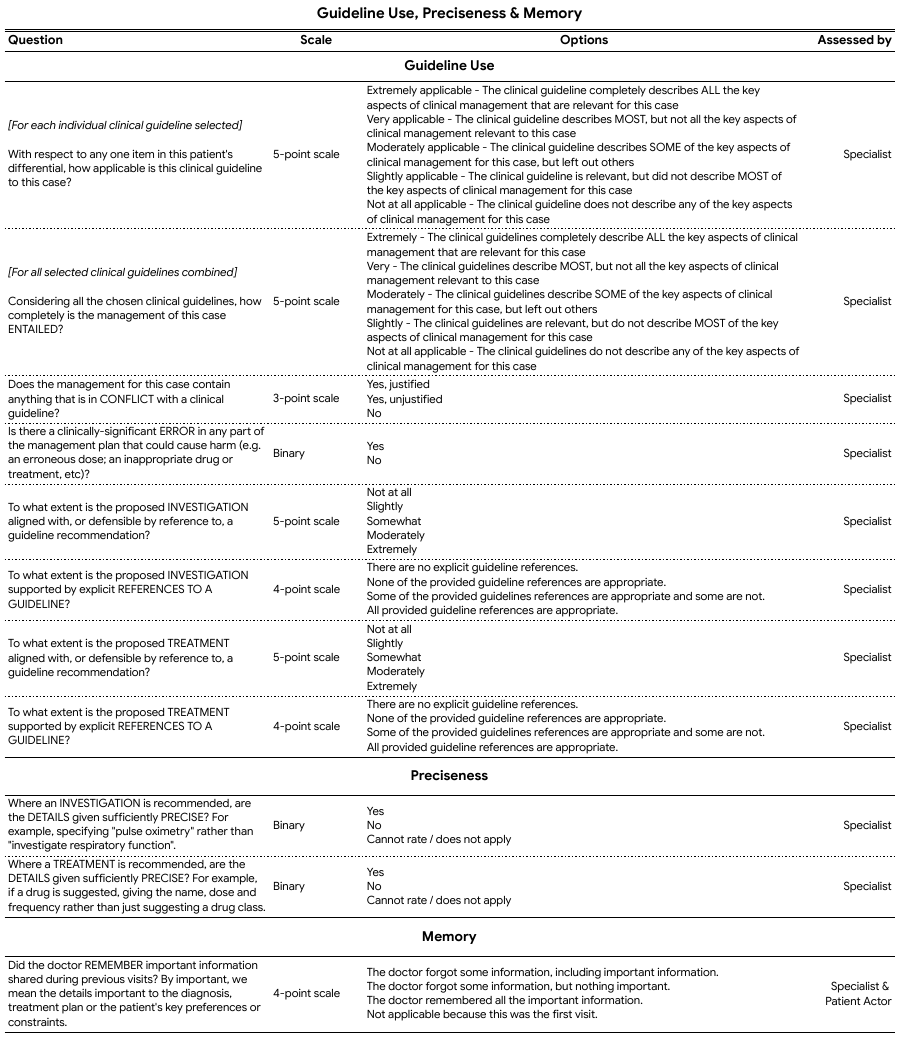}
    \caption{\textbf{Guideline Use, Preciseness and Memory rubric details.}}
    \label{tab:management_extended_rubric_details}
\end{table}
\clearpage
\section{Rating Interface for Specialist Physicians}\label{appendix:eval_ui}

\begin{figure}[hbt!]
    \centering
    \includegraphics[width=0.97\textwidth,height=\textheight,keepaspectratio]{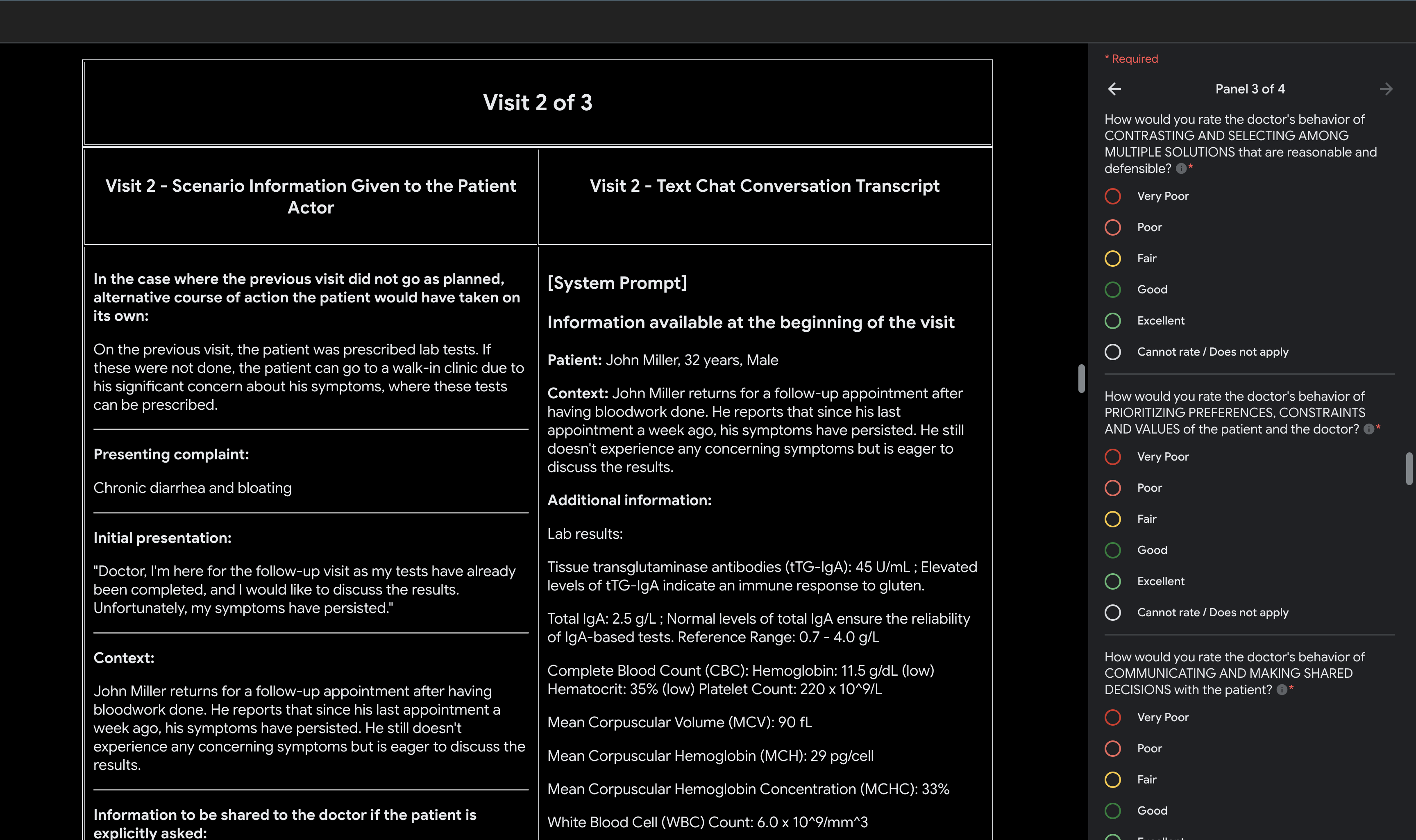}
    \vspace{0.3cm}
    \caption{Specialist physician reviews the conversation transcript for visit 2 of 3 and assesses the doctor's behavior.}
    \label{fig:specialist_evaluation_interface_a}
\end{figure}

\begin{figure}[hbt!]
    \centering
    \includegraphics[width=0.97\textwidth,height=\textheight,keepaspectratio]{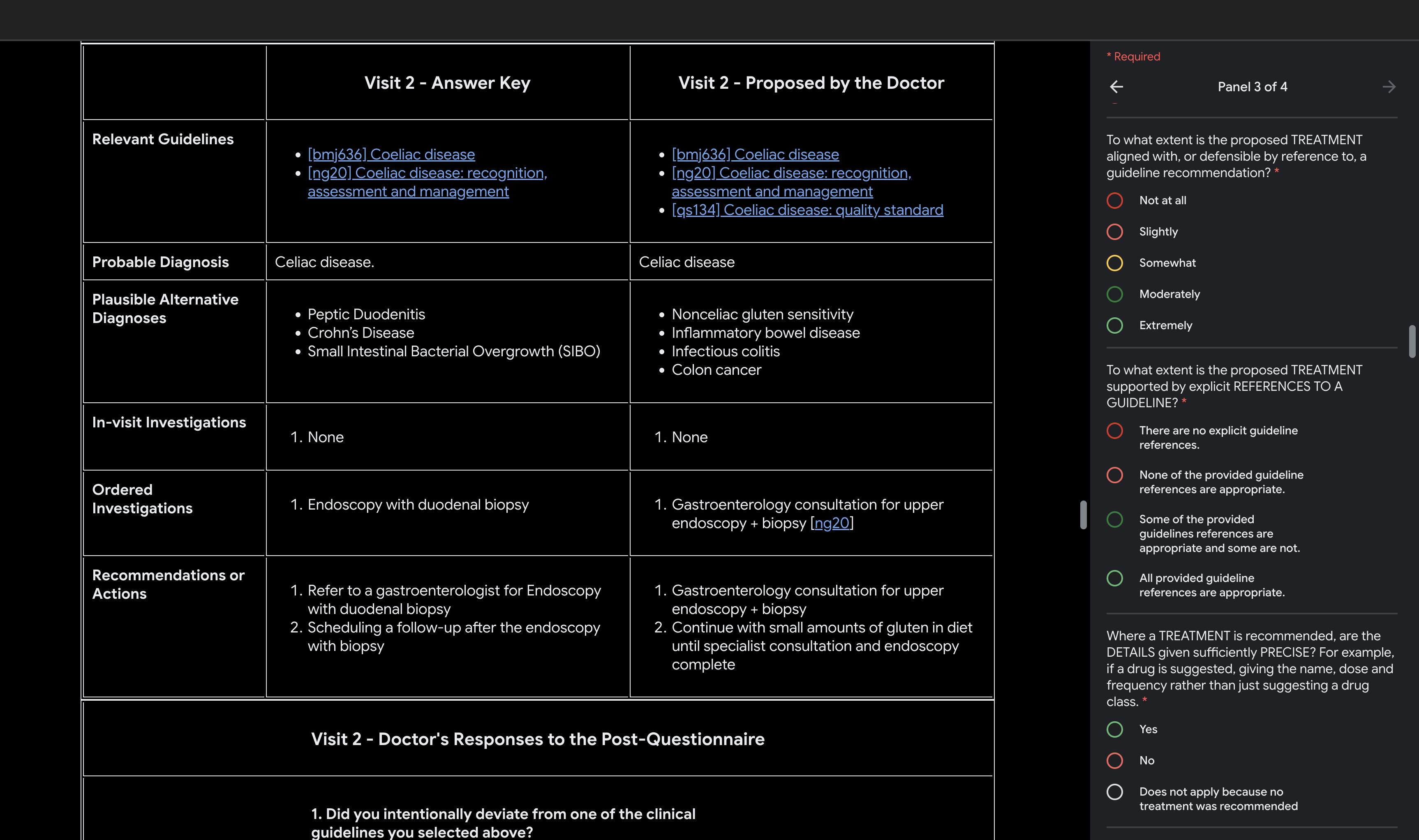}
    \vspace{0.3cm}
    \caption{Specialist physician assesses the proposed management plan and selected guidelines for visit 2 of 3.}
    \label{fig:specialist_evaluation_interface_b}
\end{figure}
\clearpage
\section{Distribution of Patient Actor Ratings}
\label{sec:distribution_of_patient_actor_ratings}

\begin{figure}[hbtp]
    \centering
    \includegraphics[width=\textwidth,height=\textheight,keepaspectratio]{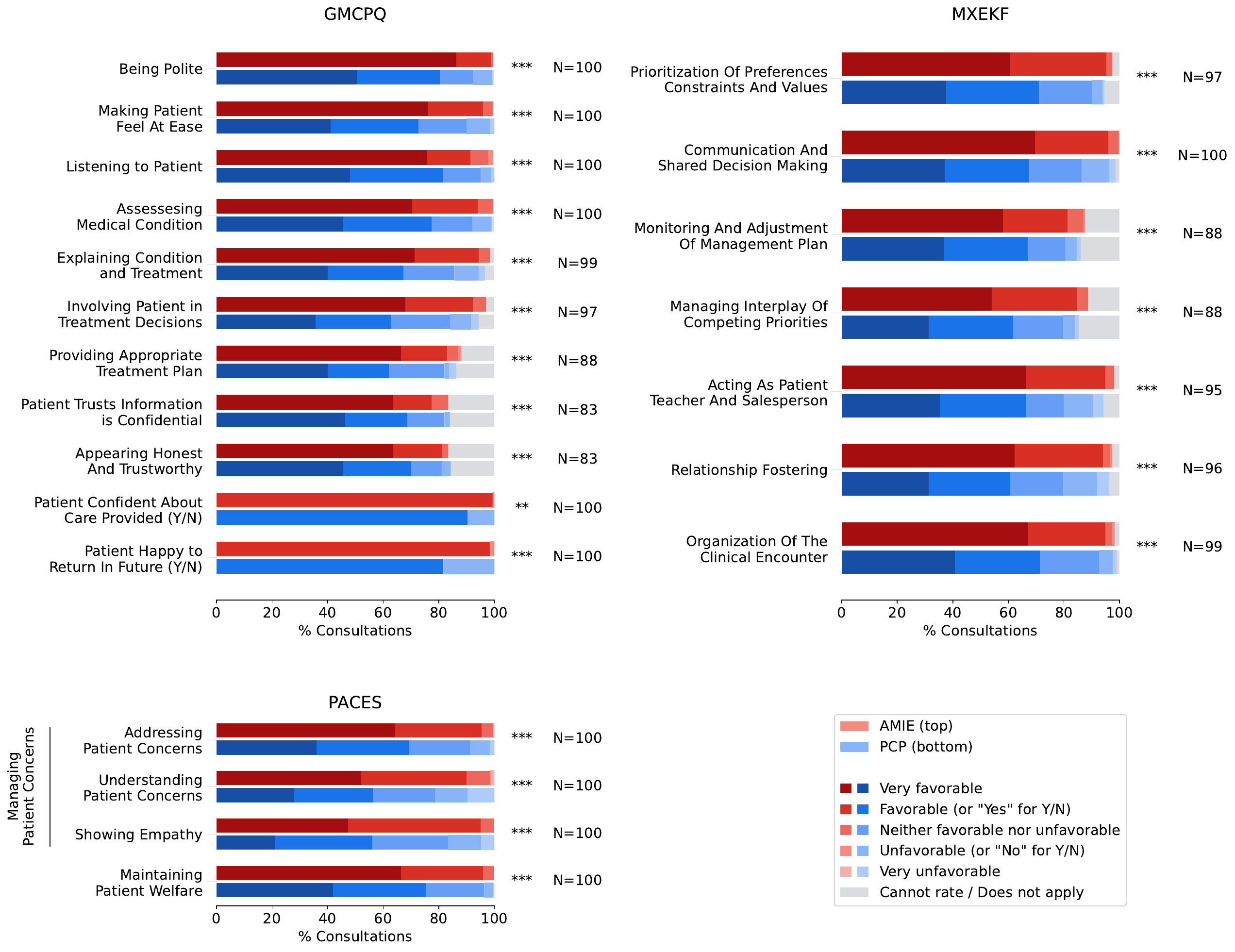}
    \vspace{0.5cm}
    \caption{\textbf{Patient actor ratings.} Conversation qualities as assessed by patient actors upon conclusion of the consultation. For illustration purposes, all responses from five-point rating scales were mapped to a generic five-point scale ranging from `Very favorable' to `Very unfavorable'. For Yes/No questions, a (positive) `Yes' response was mapped to the same color as `Favorable' and a (negative) `No' response to the same color as `Unfavorable'. Rating scales were adapted from the General Medical Council Patient Questionnaire (GMCPQ), the Practical Assessment of Clinical Examination Skills (PACES), and a set of Management Reasoning Empirical Key Features (MXEKF). Details on question wording and response options are provided in~\cref{appendix:rubrics}. Asterisks represent statistical significance ($*:p<0.05$, $**:p<0.01$, $***:p<0.001$, $n.s.: $ not significant).}
    \label{fig:patient_actor_ratings}
\end{figure}
\clearpage
\section{Distribution of Specialist Physician Ratings}
\label{sec:distribution_of_specialist_physician_ratings}

\begin{figure}[hbtp]
    \centering
    \includegraphics[width=0.65\textwidth, height=\textheight, keepaspectratio]{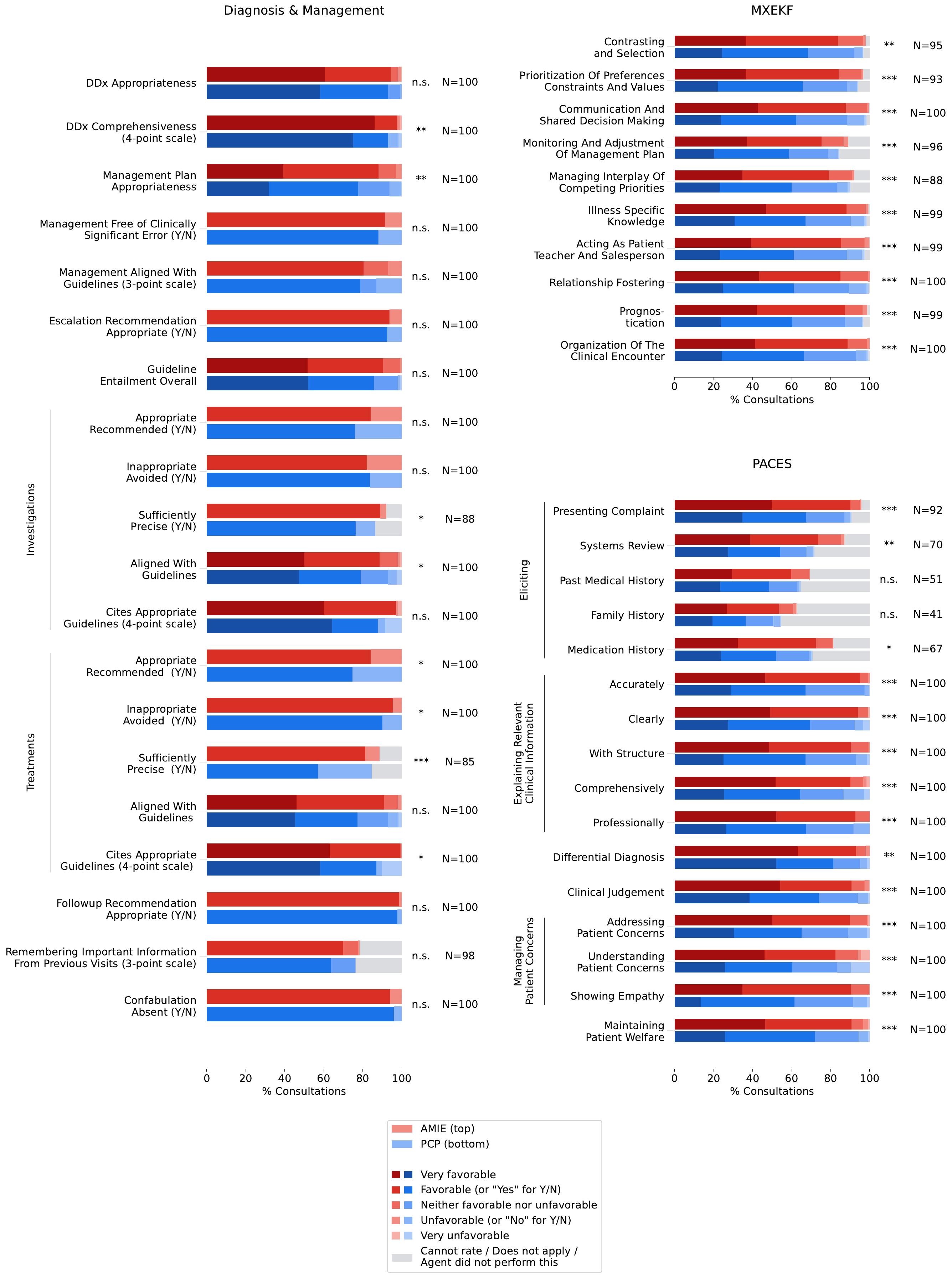}
    \vspace{0.1cm}
    \caption{\textbf{Specialist physician ratings.} Conversation and reasoning qualities as assessed by specialist physicians. For illustration purposes, all responses from five-point rating scales were mapped to a generic five-point scale ranging from `Very favorable' to `Very unfavorable'. The only four-point scale (DDx Comprehensiveness) was mapped to the same scale, ignoring the `Neither favorable nor unfavorable' option. For Yes/No questions, a (positive) `Yes' response was mapped to the same color as `Favorable' and a (negative) `No' response to the same color as `Unfavorable'. Rating scales were adapted from the Practical Assessment of Clinical Examination Skills (PACES), a set of Management Reasoning Empirical Key Features (MXEKF), and other sources (see \cref{appendix:rubrics}). Ratings from three distinct specialist physician raters for each case were aggregated using the median. Asterisks represent statistical significance ($*:p<0.05$, $**:p<0.01$, $***:p<0.001$, $n.s.: $ not significant).}
    \label{fig:specialist_ratings}
\end{figure}

\clearpage
\section{RxQA Medication Reasoning Benchmark} \label{appendix:rxqa_results}

\begin{figure}[hbt!]
    \centering
    \includegraphics[width=\textwidth,height=\textheight,keepaspectratio]{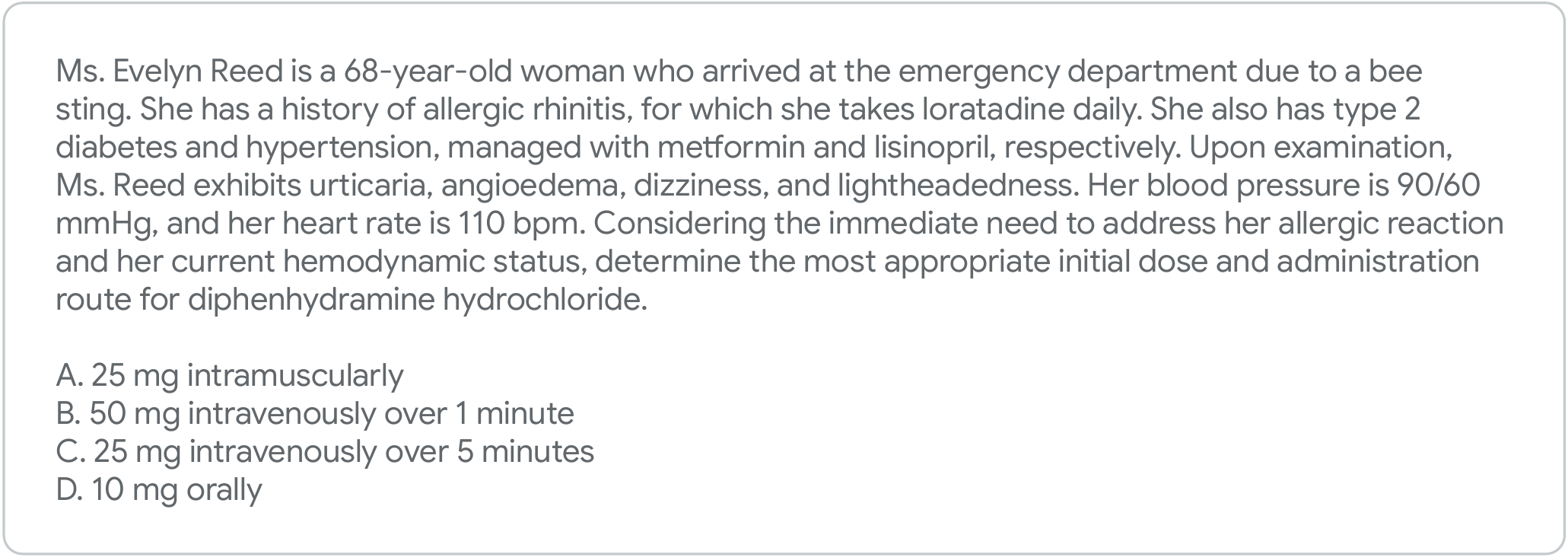}
    \vspace{0.2cm}
    \caption{\textbf{RxQA Example Question.} An example RxQA question derived from OpenFDA medication labels.}
    \label{fig:rxqa_example_question}
\end{figure}

\begin{table}[h!]
\caption{
\textbf{RxQA Medication Reasoning Accuracy.}
Comparison of question-answering accuracy between combinations of test takers (PCP, AMIE) and settings (`closed'-book, `open'-book) and for different subsets of questions.
Accuracy is shown as mean and 95\% confidence intervals for a binomial proportions.
P-values are from McNemar tests with false discovery rate (FDR) correction.
Significance levels correspond to p<0.001 (***), p<0.01 (**), p<0.05 (*) and not significant (n.s.).
Note that overlapping confidence intervals can still allow for p<0.05 as McNemar test makes use of question-level pairing.
\\
}
\label{tab:rxqa_accuracy_stats}
\begin{tabular}{p{1.7cm}p{0.5cm}p{2cm}p{2.4cm}p{2cm}p{2.4cm}p{1.3cm}p{0.7cm}}
\toprule
Pharmacist-rated Difficulty & N & Test Taker \& Setting A & Accuracy A [\%] & Test Taker \& Setting B & Accuracy B [\%] & P-value & Sig. Level \\
\midrule
\multirow[t]{6}{*}{Both} & \multirow[t]{6}{*}{600} & AMIE closed & 51.7 (47.7, 55.7) & AMIE open & 65.3 (61.5, 69.1) & 6.72e-11 & *** \\
 &  & AMIE closed & 51.7 (47.7, 55.7) & PCP closed & 43.8 (39.9, 47.8) & 5.68e-03 & ** \\
 &  & AMIE closed & 51.7 (47.7, 55.7) & PCP open & 57.0 (53.0, 61.0) & 5.53e-02 & n.s. \\
 &  & AMIE open & 65.3 (61.5, 69.1) & PCP closed & 43.8 (39.9, 47.8) & 2.40e-15 & *** \\
 &  & AMIE open & 65.3 (61.5, 69.1) & PCP open & 57.0 (53.0, 61.0) & 6.31e-04 & *** \\
 &  & PCP closed & 43.8 (39.9, 47.8) & PCP open & 57.0 (53.0, 61.0) & 5.97e-09 & *** \\
\midrule
\multirow[t]{6}{*}{Higher} & \multirow[t]{6}{*}{318} & AMIE closed & 50.6 (45.1, 56.1) & AMIE open & 57.9 (52.4, 63.3) & 1.01e-02 & * \\
 &  & AMIE closed & 50.6 (45.1, 56.1) & PCP closed & 41.5 (36.1, 46.9) & 1.29e-02 & * \\
 &  & AMIE closed & 50.6 (45.1, 56.1) & PCP open & 47.8 (42.3, 53.3) & 4.56e-01 & n.s. \\
 &  & AMIE open & 57.9 (52.4, 63.3) & PCP closed & 41.5 (36.1, 46.9) & 7.17e-06 & *** \\
 &  & AMIE open & 57.9 (52.4, 63.3) & PCP open & 47.8 (42.3, 53.3) & 3.22e-03 & ** \\
 &  & PCP closed & 41.5 (36.1, 46.9) & PCP open & 47.8 (42.3, 53.3) & 4.53e-02 & * \\
\midrule
\multirow[t]{6}{*}{Lower} & \multirow[t]{6}{*}{282} & AMIE closed & 52.8 (47.0, 58.7) & AMIE open & 73.8 (68.6, 78.9) & 1.08e-10 & *** \\
 &  & AMIE closed & 52.8 (47.0, 58.7) & PCP closed & 46.5 (40.6, 52.3) & 1.47e-01 & n.s. \\
 &  & AMIE closed & 52.8 (47.0, 58.7) & PCP open & 67.4 (61.9, 72.8) & 6.94e-04 & *** \\
 &  & AMIE open & 73.8 (68.6, 78.9) & PCP closed & 46.5 (40.6, 52.3) & 4.37e-11 & *** \\
 &  & AMIE open & 73.8 (68.6, 78.9) & PCP open & 67.4 (61.9, 72.8) & 7.08e-02 & n.s. \\
 &  & PCP closed & 46.5 (40.6, 52.3) & PCP open & 67.4 (61.9, 72.8) & 1.60e-09 & *** \\
\bottomrule
\end{tabular}
\end{table}
\clearpage
\section{RxQA Details}\label{rxqa_prompts}

Medication labels from OpenFDA and British National Formulary (BNF) included information about the medication, its brand/generic names, form, dosages, indications, contraindications, use in specific populations, interactions and more. From OpenFDA, we selected 300 commonly used drugs (see the ClinCalc DrugStats Database \cite{clincalc2022website}), as well as a random set of 500 additional drugs. 1748 drugs from BNF were used; we expected the resulting questions to be more difficult on average due to being less skewed towards well-known medications. Gemini 1.5 Flash was used to generate, validate, and filter multiple choice questions for RxQA.

\subsection{Short Question Generation}\label{easy_question_generation_prompt}

Short questions were generated zero-shot by simply providing the formatted medication label as context and asking the model to produce multiple choice questions with four options. The specific prompt was as follows:

\begin{figure}[h!]
\begin{promptbox2}{RxQA Short Question Generation}
\footnotesize
\input{prompts/rxqa_easy_question_generation.tex}
\end{promptbox2}
\vspace{0.1cm}
\caption{\textbf{Prompt:} RxQA Short MCQ Generation}
\label{fig:rxqa_short_question_generation}
\end{figure}

\subsection{Long Question Generation}\label{hard_question_generation_prompt}

Longer questions were generated through a multi-step process. For the OpenFDA-sourced questions, up to 2 related medication labels were appended to the chosen medication label, randomly selected from drug names present in the original label. We then generated a list of challenges associated with each drug, using the prompt in \cref{fig:rxqa_long_question_challenges}.
Next, we used the prompt in \cref{fig:rxqa_long_question_scenarios} to generate specific patient scenarios which test understanding of a subset of these challenges. Finally, we generated multiple choice questions based on these scenarios, each with a single correct answer using the prompt in \cref{fig:rxqa_long_question_generation}.

\begin{figure}[h!]
\begin{promptbox2}{RxQA Long Question Generation - Medication Challenges}
\footnotesize
\input{prompts/rxqa_hard_question_challenges.tex}
\end{promptbox2}
\vspace{0.1cm}
\caption{\textbf{Prompt:} RxQA Long Question Generation - Medication Challenges}
\label{fig:rxqa_long_question_challenges}
\end{figure}

\begin{figure}[h!]
\begin{promptbox2}{RxQA Long Question Generation - Patient Scenarios}
\footnotesize
\input{prompts/rxqa_hard_question_scenario.tex}
\end{promptbox2}
\vspace{0.1cm}
\caption{\textbf{Prompt:} RxQA Long Question Generation - Patient Scenarios}
\label{fig:rxqa_long_question_scenarios}
\end{figure}

\begin{figure}[h!]
\begin{promptbox2}{RxQA Long Question Generation}
\footnotesize
\input{prompts/rxqa_hard_question_generation.tex}
\end{promptbox2}
\vspace{0.1cm}
\caption{\textbf{Prompt:} RxQA Long Question Generation}
\label{fig:rxqa_long_question_generation}
\end{figure}

\clearpage
\subsection{Validation Prompts}\label{rxqa_mcq_validation_prompt}

We auto-evaluated these generated questions to help ensure they met an adequate quality standard. First, we checked whether the question itself was worded clearly and was adequately specified using the prompt in \cref{fig:rxqa_val_question_quality}.
We also confirmed with string matching that the correct answer choice was listed as one of the options. We then asked the model if it believed the listed answer was correct for the given question using the prompt in \cref{fig:rxqa_val_answer_correctness}. Finally, we asked the model if the correct answer was the \emph{only} correct answer using the prompt in \cref{fig:rxqa_val_answer_uniqueness}, removing questions which had multiple good answer choices.

\begin{figure}[h!]
\begin{promptbox2}{MCQ Validation - Question Quality}
\footnotesize
\input{prompts/rxqa_validation_question_quality.tex}
\end{promptbox2}
\vspace{0.1cm}
\caption{\textbf{Prompt:} MCQ Validation - Question Quality}
\label{fig:rxqa_val_question_quality}
\end{figure}

\begin{figure}[h!]
\begin{promptbox2}{MCQ Validation - Answer Correctness}
\footnotesize
\input{prompts/rxqa_validation_answer_correctness.tex}
\end{promptbox2}
\vspace{0.1cm}
\caption{\textbf{Prompt:} MCQ Validation - Answer Correctness}
\label{fig:rxqa_val_answer_correctness}
\end{figure}

\begin{figure}[h!]
\begin{promptbox2}{MCQ Validation - Answer Uniqueness}
\footnotesize
\input{prompts/rxqa_validation_answer_uniqueness.tex}
\end{promptbox2}
\vspace{0.1cm}
\caption{\textbf{Prompt:} MCQ Validation - Answer Uniqueness}
\label{fig:rxqa_val_answer_uniqueness}
\end{figure}

\subsection{MCQ Selection}\label{rxqa_mcq_filtering_prompt}
Gemini 1.5 Flash was asked to answer each generated question zero-shot, both with and without the medication context. We identified the set of questions which the model answered correctly when it had the context, and incorrectly without the context. Note that what was considered correct at this stage was not necessarily the same as what the pharmacists selected to be correct in the next step (\cref{rxqa_pharmacist_revision}). For both the OpenFDA-sourced and BNF-sourced questions, we randomly sampled 100 short and 200 long questions which met this criteria as well as the validation criteria in \cref{rxqa_mcq_validation_prompt} to send out for pharmacist revision.

\subsection{Pharmacist Revision}\label{rxqa_pharmacist_revision}
The questions and selected answer options were each revised by one of four pharmacists in each jurisdiction. In revision, the pharmacist was allowed to change the wording of the question and all of answer choices to ensure that the question was high quality and had exactly one correct answer. Following this revision, the pharmacist selected what they believed to be the correct answer, which may not be the same as what the model initially believed. Afterwards, the pharmacists also rated the difficulty of each question on the following scale: Trivial, Easy, Medium, Hard, and Impossible. For subgroup analysis purposes, Trivial and Easy were mapped to lower difficulty, while Medium, Hard, and Impossible mapped to higher difficulty. These pharmacist-revised set of questions are what make up the final RxQA benchmark. An example question, derived from OpenFDA, is shown in \cref{fig:rxqa_example_question}.


\setlength\bibitemsep{3pt}
\printbibliography
\balance
\clearpage
\end{refsection}

\end{document}